\documentclass[%
nohebmain, % English is the main language
nothanks,
twoside,openany,12pt,
nohebibm,nostdtheo,encutf8
]{hebtech}

\usepackage{amsmath}
\usepackage{amssymb}
\usepackage{ntheorem}

\usepackage[stable,perpage,symbol*,multiple]{footmisc}
\usepackage[tracking=true,kerning=true,spacing=true]{microtype}

\usepackage{float}
\usepackage{subfig}
\usepackage{ifpdf}

\ifpdf
\usepackage[final,pdftex]{graphicx}
\usepackage[pdftex,dvipsnames,usenames]{color}
\DeclareGraphicsExtensions{.pdf}
\else
\usepackage[final,dvips]{graphicx}
\usepackage[dvips,dvipsnames,usenames]{color}
\DeclareGraphicsExtensions{.eps}
\fi

\usepackage{chicago}

\theoremstyle{break}
\newtheorem{Lemma}{Lemma}
\newtheorem{Theorem}{Theorem}

\theoremstyle{plain}
\theoremheaderfont{\sc}
\theorembodyfont{\upshape}
\theoremstyle{nonumberplain} \theoremseparator{}
\theoremsymbol{\rule{1ex}{2ex}}
\newtheorem{Proof}{Proof:}

\author{% Hebrew name:
}{% English name:
  Eliyahu Osherovich
}

\title{% In Hebrew
}{% In English
  Ant Robotics: Covering Continuous Domains by Multi-A(ge)nt Systems
}

\dept{%Hebrew
}{% English
  Department of Computer Science%
}

\dom{% Hebrew
}{% English
  Computer Science%
}

\advisori{% Hebrew
}{% English
  Prof.~Alfred~M.~Bruckstein
}
\date{15}{12}{2006}% just a date..

\losymtext{
  \begin{tabular*}{\columnwidth}%
    {p{0.25\columnwidth}@{}p{0.7\columnwidth}}
    $\mathbf{\Omega}$ &bounded connected domain\\
    \\
    $\mathbf{\|a-b\|}$ &geodesic distance between points $\mathbf{a}$ and
    $\mathbf{b}$\\
    \\
    $\mathbf{d}$ &diameter of $\mathbf{\Omega}$ (length of the ``longest
    shortest path'' in $\mathbf{\Omega}$)\\
    \\
    $\mathbf{\sigma(a,t)}$ &the marker value of point $\mathbf{a}$ at
    time instance  $\mathbf{t}$.\\
    \\
    $\mathbf{D(r,c)}$  &open disk of radius $\mathbf{r}$ centered at
    $\mathbf{c}$ : $\mathbf{\left\{p\in\Omega~\Big\vert~\|p-c\|< r
      \right\}}$\\
    \\
    $\mathbf{R(r_1,r_2,c)}$  &ring with radii $\mathbf{r_1}$ and $\mathbf{r_2}$
    centered at  $\mathbf{c}$ :
    $\mathbf{\Bigl\{p\in\Omega ~\Big\vert~ r_1\leq \|p-c\|\leq
      r_2 \Bigr\}}$\\
  \end{tabular*}
}

\abstracttext{
  In this work we present an algorithm for covering continuous connected
  domains by ant-like robots with very limited capabilities. The
  robots can mark  visited
  places with pheromone marks and sense the level of the pheromone  in
  their local neighborhood.
  In case of multiple robots these pheromone marks can be sensed by all robots and
  provide the only way of (indirect) communication between the robots. 
  The robots are assumed to be
  memoryless, and to have no global information such as the domain
  map, their own  position (either absolute or relative), total marked
  area percentage, maximal pheromone level,
  etc.. Despite the robots' simplicity, we 
  show that they are able, by  running a very simple  rule of behavior,
  to ensure efficient covering of  arbitrary connected
  domains, including non-planar and multidimensional ones. The novelty of our
  algorithm lies in the fact that, unlike previously proposed methods,
  our algorithm 
  works on  continuous domains without relying on some ``induced''
  underlying graph, that effectively reduces the problem to a discrete
  case of graph covering. The algorithm guarantees
  complete coverage of any connected domain. We also prove that the
  algorithm is noise immune, i.e., it is able to cope with any initial
  pheromone profile (noise). In addition the algorithm provides a
  bounded constant time between two successive visits of the robot,
  and thus, is suitable for patrolling or surveillance applications.
}

\graphicspath{{figs/}{running_figs/}{figs/asy/}}

\begin{document}
\thesisheader

\chapter{Introduction}
\section{Motivation}
\label{sec:introduction}
Suppose we want to cover (or clean or search or paint) a connected domain
in $\mathbb{R}^2$ with one or more simple robots that have an effector
(or arm) that can sweep a well-defined neighborhood of the robots when
they are stationary.
We shall say that a domain was \emph{covered} by the (team of) robots
if each and every 
point of the domain was swept by a robot effector. In fact, every
time we want to build an automatic machine suitable for applications
such as floor cleaning, snow removal, lawn mowing, painting, mine-field
de-mining, unknown terrain exploration and so forth, we face the
problem of complete covering of  domains by such devices.

\section{Problem Constraints}
\label{sec:constraints}
The approach to solving covering problems depends, of course, on
the capabilities of our robots and on various environmental
constraints. Hence many algorithms can be, and actually
have been, developed to accommodate constraints and assumptions on the
robots used for the
covering problem. The various considerations
are:
\begin{enumerate}
\item The domain type (e.g., discrete versus continuous, a simply
  connected or multiply connected region, a general graph or a grid, etc.)
\item The capabilities of robots (their communication means, the
  amount of on-board memory, the size of footprint and the areas swept
  by the effectors and range of robots' sensors)
\item The type of knowledge the robots are assumed to be able to get
  or gather via their sensors
  (either global or local, often referred as off-line and
  on-line operation respectively)
\item The local behavior and interaction model in case of multiple
  robots (such as synchronous or asynchronous operation, centralized
  or  distributed control, etc.)
\end{enumerate}

In this paper we adopt the model proposed in~\cite{yanovski01vertex}, which
assumes that the robots are anonymous, (i.e.,
all robots are identical), memoryless, (i.e.,  have no ability to ``remember''
anything from the past), and have no means of direct communication which
means there is no direct exchange of information between the robots.
In fact, our robots are (most of the time) completely unaware of the
existence of other robots and their only means of (indirect)
communication is via some marks they leave in their environment.
This model was originally inspired by ants and other
insects using chemicals called \emph{pheromones} that are left on
the ground as a mean of achieving indirect communication and
coordination.\footnote{It is fair to say that recent publications
  show that ants are not memoryless (see, e.g.,~\cite{harris05ant}).
  However it is certainly  true that for certain ant species
  these individual capabilities play a limited role in navigation and
  trail laying or trail following mechanisms.}
Ant colonies, despite
the simplicity  of 
single ants, demonstrate surprisingly good results in global problem
solving and pattern
formation~\cite{bruckstein93why,hoelldobler90ants,schoene84spatial,dorigo96ant,dorigo99ant}. 
  Consequently,
ideas borrowed from  insects behavior research are becoming increasingly popular
in ant-robotics and distributed
systems~\cite{dorigo96ant,dorigo99ant,wagner01from,bonabeau00swarm,russell99ant,koenig01terrain}.
Simple robots were found to be capable of performing quite complex distributed
tasks while providing the benefits of being small, cheap, easy to produce
and easy to maintain.

This thesis is organized as
follows. Our formal robot model is presented in
Section~\ref{sec:robot-model}. In Section~\ref{sec:mark-ant-walk} we
define the Mark-And-Walk (MAW) covering algorithm, followed by a
short survey of previously proposed covering algorithms in
Section~\ref{cha:related-work}. As mentioned earlier the number of
such algorithms is fairly large, therefore, we limited our survey to those
that share some common principles with the algorithm suggested in this
paper. Formal proofs of  complete coverage and efficiency analysis are
given in Sections~\ref{sec:maw-proof-coverage}
and~\ref{sec:maw-effic-analys} respectively. In
Section~\ref{sec:extensions} we provide various extensions 
of the basic MAW algorithms including their applicability for
multi-robot systems and
their performance when the environment contains noise (false pheromone
marks). Results of simulations
are presented 
in Section~\ref{sec:simul-exper}. Section~\ref{sec:conclusions}
provides a summary of our results and a discussion of possible
extensions and implementation details.
% Unanswered questions and topics for
% future research are given in Section~\ref{sec:future-work}.

\chapter{The Mark-Ant-Walk (MAW) Covering Algorithm}
\label{cha:mark-walk-covering}
\section{Robot Model}
\label{sec:robot-model}
Below we define the mathematical problem of robot covering along with
the model for the robots that we use throughout
this paper.

The domain to be covered will be denoted by $\Omega$. At the moment 
we consider only  two-dimensional domains (however, extensions will be
given in Section~\ref{sec:non-eucl-metr}). Given any two points
$a,b\in\Omega$  denote the \emph{geodesic distance} between $a$ and $b$ as
$\|a-b\|$, i.e., the length of the shortest path that connects $a$ and
$b$, restricted to lie entirely in the domain $\Omega$. For the sake
of brevity we shall  omit the word ``geodesic'' and use simply
``distance''. At the moment, we assume that this length is measured as
a common Euclidean length in two-dimensional space; extensions to other
measures being discussed  in Section~\ref{sec:non-eucl-metr}.
We say that a robot is located at point $p \in \Omega$ if the
``center'' of the robot lies at $p$. We shall then assume that the
robot is able to sense the pheromone level at 
its current position $p$ and in a closed ring $R(r, 2r, p)$ lying
between the  internal radius
$r$ and the external radius $2r$
around $p$. $R$ is formally defined as follows:
\begin{equation}
  \label{eq:1}
  R(r_1, r_2, c) =
  \left\{
    a \in\Omega\ \Big\vert\ r_1 \leq \|a-c\| \leq r_2
  \right\},
\end{equation}
where $r$ is  considered to be an intrinsic parameter of the
robot.
% \footnote{ It is convenient to think 
%that the robot has a circular footprint of radius $2r$ and a
%concentric circular effector of radius $r$. Note also that whenever
%we speak about the robot's location we mean the location of its
%center.} 

Additionally, our robot is able to set a constant arbitrary pheromone
level in the area swept by it effector, which is, we assume, an
open disk $D(r, p)$ of radius $r$ around its current location $p$. The
formal definition of  $D$ is as follows:
\begin{equation}
  \label{eq:2}
  D(r,c) =
  \left\{
    a \in \Omega\ \Big\vert\ \|a-c\|<r
  \right\}.
\end{equation}

\begin{figure}[H]
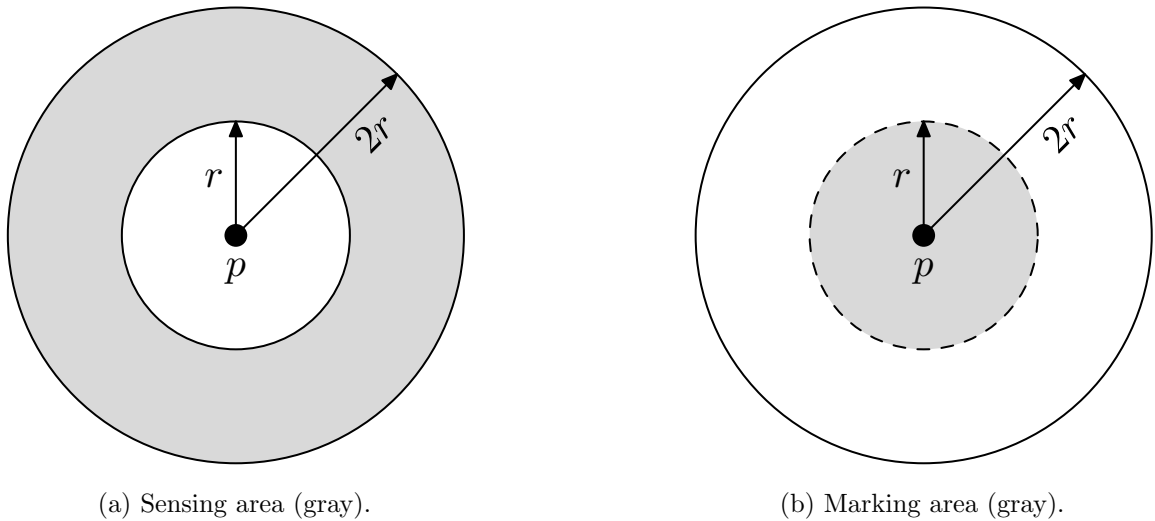

  \centering
  \subfloat[Sensing area (gray).]{\includegraphics[width=0.4\textwidth]{maw.1}}
  \hfill
  \subfloat[Marking area (gray).]{\includegraphics[width=0.4\textwidth]{maw.2}}
  \caption{Robot's sensing and marking areas}
  \label{fig:capabilities}
\end{figure}
\begin{figure}[H]
  \centering
  \includegraphics[width=\textwidth]{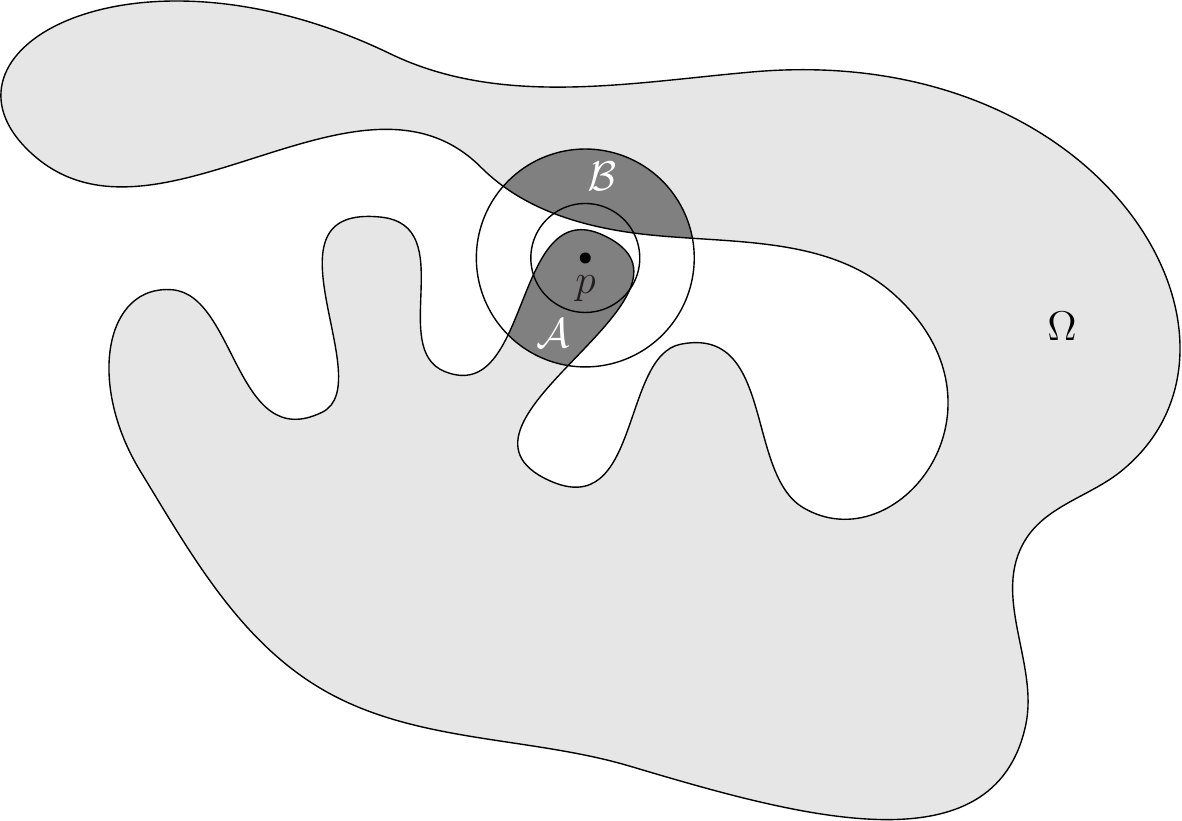}
  \caption{Geodesic distance example}
  \label{fig:geodesics}
\end{figure}

Note that, as we mentioned before, all distances are measured as
geodesic ones, hence, for example, only area $\mathcal{A}$ in
Figure~\ref{fig:geodesics} will be available to the robot. Area
$\mathcal B$, on the other hand, will not be ``visible'' to the robot
since the distance from the robot location $p$ to any point in
$\mathcal{B}$ is greater than $2r$.

Furthermore, we shall assume that our time steps are discrete. We denote by
$\sigma(a,t)$ the pheromone level of point $a\in\Omega$ at time
instance $t$, where $t = 0, 1, 2, 3, \ldots$.
\section{The Mark-Ant-Walk (MAW) Algorithm}
\label{sec:mark-ant-walk}
Initially, we consider the case where no point is marked with the pheromone, thus all $\sigma$
values are assumed to be equal to zero:
\begin{equation}
  \label{eq:3}
  \sigma(a,0) = 0;\ \forall a\in\Omega.
\end{equation}
Suppose further that a starting point is (randomly) chosen for a (single) robot and then a MAW
step rule (as described in Table~\ref{tab:MAW_Rule}) is
applied repeatedly. Note that there is no explicit stopping 
condition for this algorithm; nevertheless, one can use an upper bound
(that will be 
provided later in this paper) on the cover time in order to stop robots after a
sufficient time period has elapsed that effectively guarantees complete covering.
\begin{figure}[H]
  \centering
  \begin{tabular}{|p{0.1\columnwidth}@{\hspace{0pt}}p{0.8\columnwidth}|}
    \hline
    \multicolumn{2}{|p{\columnwidth}|}{{\bf Mark-Ant-Walk step rule} (current time
      is $t$ and robot location is $p$)}\\
    \hline
    (1) & Find $x$ := a point in $R(r,2r,p)$ with \emph{minimal} value of $\sigma(x,t)$\\
    & (In case of a tie, i.e., when the minimal value achieved at several places - make an arbitrary decision\\ 
    & {\it /* note that $\|p-x\|\geq r$ */}\\
    (2) & If $\sigma(p) \leq \sigma(x)$ then $\forall u \in D(r,p)$
    set $\sigma(u)=\sigma(x)+1$\\
    & {\it /* we mark open disk of radius $r$ around current location */}\\
    (3) & $t := t+1$\\
    (4) & move to $x$\\
    \hline
  \end{tabular}
  \caption{MAW step rule}
  \label{tab:MAW_Rule}
\end{figure}
It is easy to see that the robot markings create some kind of \emph{potential field},
where high values of the pheromone level roughly indicate areas
that have been visited many times up to the moment and lower values of
the pheromone level correspond to a smaller number of robot's
visits (we say ``roughly indicate'' because there is no strict
  correspondence between the pheromone level and the number of the
  robot visits, since marking (setting a new pheromone level) does not
  necessarily happen in every step). According to the MAW rule, the
robot tries to avoid areas with high pheromone values by moving toward lower
levels, striving to reach areas yet uncovered.

\chapter{Related Work}
\label{cha:related-work}
In this chapter we shall discuss previous work that is close  to
our paradigm, i.e., we shall review the history of the
``pheromone-powered'' algorithms. 

A step toward a ``pheromone-marking''-oriented model was taken 
by Blum and co-workers ~\cite{blum77capability,blum78power} where \emph{pebbles} were used
to assist the search. Pebbles are tokens that can be placed on the
ground and later removed. The idea of using pebbles for unknown graph
exploration and mapping  was further
developed in \cite{bender98power}.
\section{Discrete Domains Covering Problem}
\label{sec:dicr-cove-probl}
Covering of discrete domains (graphs) is an important problem
theoretical computer science  and thus
a number of solutions  have been
proposed and comprehensively analyzed by researchers.  
Well known examples are the Breadth-First Search (BFS)  and
the Depth-First Search (DFS) algorithms for graph
traversal. Both algorithms provide excellent results in terms of
time complexity. Formal proofs of complete coverage and efficiency
analysis can be found in  books on discrete mathematics and
algorithms (see, e.g.,~\cite{cormen90introduction}). 
Note  that the DFS algorithm can 
be readily adapted to fit our robot model  as opposed to the BFS
algorithm
which requires
additional on-board memory in order to maintain a queue of already
discovered, but yet 
unvisited vertices. Moreover, the amount of on-board memory required
depends on the graph to be explored.

Two further algorithms that fit
our paradigm entirely, i.e. they relay on a group of identical
autonomous robots that mark the ground 
with pheromones, were suggested for efficient and robust graph
covering. One, called the Edge-Ant-Walk, marks the graph 
edges~\cite{wagner96smell}. Another one,  called the Vertex-Ant-Walk,
leaves
marks on graph vertices instead~\cite{yanovski01vertex,wagner98efficiently}. Both
algorithms provided significant improvement over DFS in terms of robustness
along with quite efficient cover times. Both  are capable of
completing the traversal of the graph in multi-robot cases even in the
case when almost all 
robots die and/or the environment graph changes (edges/vertices are
added or deleted) during the execution.

\section{Continous Domains Covering Problem}
\label{sec:cont-doma-cover}
Covering continuous domains is a relatively new problem. Several
algorithms addressing this problem are summarized in a good survey paper by
Choset~\cite{choset01coverage}. We shall provide below a short
review of algorithms developed for robot models that are close to the
one used in this thesis.

\subsection{Random Walk and Probabilistic Covering}
\label{sec:rand-walk-prob}
Random walks were defined for both  discrete and
continuous domains and enjoy unrivaled robustness and scalability;
however, they cannot guarantee complete coverage and can be analyzed
in terms of 
expected coverage time only. We would like to have solutions 
that guarantee complete coverage after a deterministic and bounded time period.

\subsection{Motion Planning Guided by a Potential Field}
\label{sec:moti-plann-guid}
A very popular approach is to  introduce an \emph{artificial potential field} concept
in order to accomplish the robot motion planning task
(e.g.~\cite{khatib86realtime,zelinsky93planning}).
In~\cite{zelinsky93planning}, for example,  the
authors used the distance transform as the potential field. Such an approach
could also be adopted by our robots, the potential being represented
by the odor level. However, this type of work assumed that the potential
field can be constructed prior to the 
start of robot's motion and thus requires a global knowledge of the
domain boundaries and obstacles. Such knowledge is not
available  in our model.
However, similar results could  be achieved if we assumed that obstacles were
actually the pheromone source and that the odor level decreases
gradually as we move away from them thereby forming some
kind of distance transform.

\subsection{Trail Lying Algorithms for Continuous Domains}
\label{sec:trail-lying-algor}
Several authors proposed to use trails that mark
the path traveled by the robot so far, and proposed local behavior for
the robots that resulted in some kind of 
peeling/milling of the domain.
Here two
major approaches for organizing the motion exist: contour-parallel and
direction-parallel. In the 
former approach, the robot moves along the boundaries of the domain; in
the latter one, the robot moves back and forth in some predefined
direction. These approaches often fail for non-convex domains and
thus the whole domain must be approximated as a union of convex
non-overlapping cells,
which, in turn, can be covered  with the assumed type of
motion~\cite{choset97coverage,butler00distributed,acar03path}. Another
representative of trail-based algorithms is the Mark-And-Cover (MAC)
algorithm presented in~\cite{wagner-mac}, which is actually an adaptation of the DFS
algorithm to continuous domains. This algorithm provides efficient and
effective coverage with a provable bound on cover time. Additionally, the
robot model used in the paper fits our paradigm very well.
Nevertheless, the
drawback of the MAC algorithm, and, in fact, of all trail-based algorithms,
is their sensitivity to noise and robot failure. Moreover, in
the multi-robot case trails  of the 
robots interfere with the motion of the others and may hamper their
efficiency. Another shortcoming of these 
algorithms is seen in the situation when the domain is required to be
covered repeatedly, e.g., in surveillance tasks or in the scenario
described in~\cite{gage94randomized} where autonomous robots are used
to de-mine minefields using imperfect sensors (in the sense that the probability of
a mine detection is less than 1). Our algorithm guarantees that the
whole domain is covered repeatedly, time after time. Furthermore, the
time between two successive visits at any point can be bounded in
terms of the (unknown) size of the problem (see
Section~\ref{sec:repetitive-coverage}).

\subsection{Tessalating Algorithms for Continuous Domains}
\label{sec:tess-algor-cont}
Another possible approach is to split the domain into tiles such that each
tile is easily covered by the robots (e.g., convex tiles are
suitable for ``onion peeling'' or ``back and forth''
algorithms). After a particular tile is covered, the robot
goes to a new one that is a  neighbor of  the current tile. This
approach, in fact, takes us back to a graph covering of an underlying
graph whose vertices are associated with the tiles, the
edges between vertices being defined according to the 
inter-tile connectivity.    

\chapter{The MAW Algorithm: Formal Proof and Efficiency Analysis}
\label{cha:maw-algorithm:-proof}
\section{MAW - Proof of Complete Coverage}
\label{sec:maw-proof-coverage}
Let us first show that a single robot governed by the MAW rule covers any
connected bounded domain in a finite number of steps. The outline of
the proof is as follows.

First, we prove that at any time instance, any
two points that are close enough, i.e., their distance from each other
is less than or equal to $r$, must have pheromone levels that
differ by one at most. This property closely resembles  Lipschitz
continuity for functions $f(\cdot)$:
\begin{equation}
  \label{eq:4}
  |f(p_1) - f(p_2)| \leq \kappa \|p_1 - p_2\|,
\end{equation}
for some constant $\kappa$.
However, our $\sigma$ function measuring the level of the pheromone at
each location is, by definition, not continuous and thus cannot be
Lipschitz continuous. Instead, $\sigma$ obeys the following inequality:
\begin{equation}
  \label{eq:5}
  |\sigma(p_1) - \sigma(p_2)| \leq \kappa\|p_1 - p_2\| + 1.
\end{equation}

We call this
\emph{the proximity principle}, and it  has also been 
used in several previous
papers~\cite{yanovski01vertex,wagner96smell,wagner98efficiently}. 

Second, we look
at the diameter $d$ of the domain, defined as the length of the
longest geodesic path that can be embedded in the domain: consider a pair of
points $a,b\in\Omega$. Since $\Omega$ is connected, there is at least
one path that connects $a$ and $b$ and lies entirely in the domain
$\Omega$. Among all such paths (connecting $a$ and $b$, that are
restricted to lie entirely in 
the domain $\Omega$), there is at least one that is the shortest.
We call the length of this shortest path the \emph{distance}
between $a$ and $b$ and denote it by $\|a - b\|$. Among all possible
pairs of points $a,b\in\Omega$, there exists a pair $(a_0, b_0)$ that
has the greatest possible distance between them, i.e.,
\begin{equation}
\|a_0 - b_0\| \geq \|a - b\| \; \forall a,b \in\Omega.
\end{equation}
We call the length of this longest geodesic path the \emph{diameter}
of the domain and denote it by $d$. In other words:
\begin{equation}
  \label{eq:41}
  d \triangleq \sup_{(a,b)\in\Omega}\|a - b\|.
\end{equation}

Assuming that $d$ is
finite (actually the requirement that the domain $\Omega$ is bounded
is related to its diameter and its perimeter), we easily conclude
with the aid of the proximity principle that, at any time $t$ for any two points
$a,b\in\Omega$, the difference between the pheromone levels of these
two points is upper bounded by $\lceil d/r \rceil$. This, in turn, means
that once a value of $\lceil d/r\rceil + 1$ is reached by $\sigma()$ at any point,
no unmarked point may exist and thus the whole domain has been covered by the
robot. Finally, we  show that the maximal pheromone marker value goes to
infinity as time goes to infinity hence we shall surely, at some time reach the
value of $\lceil d/r\rceil + 1$. Formal proofs are given below.

\begin{figure}[H]
  \centering
  \includegraphics[width=0.4\textwidth]{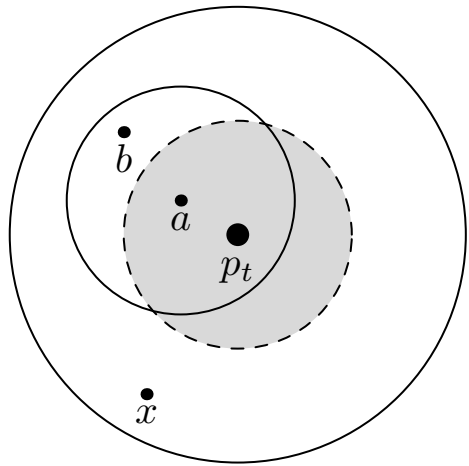}
  \caption{Close points have close marker values}
  \label{fig:proximity}
\end{figure}

\begin{Lemma}
  \label{lem:close_points}
  The difference between marker values of near by points is bounded. 
  \[
  \forall t;\forall a,b \in\Omega\text{ : if }\|a-b\| \leq r
  \text{ then } |\sigma(a,t)-\sigma(b,t)| \leq 1
  \]
\end{Lemma}
\begin{Proof}
  Note that the distance $\|a-b\|$ is the length of the shortest
  path between these two points, which is restricted to lie entirely
  in the domain $\Omega$.
  We shall prove the lemma by mathematical induction on the step
  number. The lemma is clearly true at $t=0$ when all the marks are
  assumed to be zero. Assuming it is also
  true at time $t$, we shall show it remains true at time $t+1$. Let
  us look at two arbitrary points $a,b \in \Omega$, such that
  $\|a-b\| \leq r$. In the trivial case neither $a$ nor $b$
  change their marker values at the $(n+1)$th step; therefore, the lemma
  continues to hold.
  If both $a$ and $b$ change
  their values, then $\sigma(a,t+1)=\sigma(b,t+1)$  since
  the algorithm assigns the same values to all the points it
  changes. Hence the only interesting case is when only one point (say
  $a$) changes its marker value, while the other one remains
  unchanged. Assuming the current robot's location is $p_t$ we conclude
  that $a$ belongs to $D(r, p_t)$, otherwise it could not change its marker
  value. Therefore, $\|a -p_t\| < r$. However, $b$ does not
  change its marker value and thus $\|b -p_t\| \geq r$. Combining
  these constraints  we get:
  \begin{equation}
    \label{eq:44}
    \left\{
      \begin{array}{l}
        \|a - b\| \leq r \\
        \|a - p_t\| \leq r\\
        \|b - p_t\| \geq r
      \end{array}
      \right.
      \Rightarrow
      r \leq \|b - p_t\| \leq 2r, 
  \end{equation}
  or, equivalently, $b \in R(r, 2r, p_t)$. This situation is depicted in
  Figure~\ref{fig:proximity}. Now let us recall how the new marker
  value of $a$ is determined. First, we look for the minimal marker
  value among all points in $R(r,2r, p_t)$. Assume that this value
  is attained at some point $x\in R(r,2r, p_t)$.  The new
  marker value of $a$ is then set if and only if the pheromone level
  at the current robot's location is smaller than or equal to that at
  $x$: $\sigma(p_t,t) \leq \sigma(x,t)$. In this case we have: 
  \begin{equation}
    \label{eq:changing_a}
    \sigma(a,t+1) = \sigma(x,t) + 1.
    \end{equation}
    Since both points $x$ and $b$ belong to $R(r,2r, p_t)$,
    we have
    \begin{equation}
      \sigma(b,t) \geq \sigma(x,t),
    \end{equation}
    because of the way the point $x$ was chosen.  
    Now, on the one hand, we have:
    \begin{equation}
      \label{eq:sigma_a_big}
      \left\{
        \begin{array}{l}
          |\sigma(a,t) - \sigma(b,t)| \leq 1\\
          \sigma(b,t) \geq \sigma(x,t)
        \end{array}
      \right.
      \Rightarrow
      \sigma(a,t) \geq \sigma(x,t) - 1;
    \end{equation}
    and on the other hand:
    \begin{equation}
      \label{eq:sigma_a_small}
      \left\{
        \begin{array}{l}
          |\sigma(a,t) - \sigma(p_t,t)| \leq 1\\
          \sigma(p_t,t) \leq \sigma(x,t)
        \end{array}
      \right.
      \Rightarrow
      \sigma(a,t) \leq \sigma(x,t) + 1.
    \end{equation}
    Combining inequalities (\ref{eq:sigma_a_big}) and
    (\ref{eq:sigma_a_small}), we get
    \begin{equation}
      \label{eq:sigma_a_both}
      |\sigma(a,t) - \sigma(x,t)| \leq 1.
    \end{equation}
    Using the system of inequalities~(\ref{eq:sigma_a_big}), we
    conclude that
    \begin{equation}
      \label{eq:sigma_b_big}
      0 \leq \sigma(b,t) - \sigma(x,t) \leq 2.
    \end{equation}
    Combining the above inequality with the fact that $\sigma(a,t+1) = \sigma(x,t) + 1$
    and $\sigma(b,t+1) = \sigma(b,t)$, we get the desired result:
    \begin{equation}
      \label{eq:proximity_lemma}
      |\sigma(a,t+1) - \sigma(b,t+1)| \leq 1.
    \end{equation}
    Thus the lemma is proved.
  \end{Proof}

\begin{figure}[H]
  \centering
  \includegraphics[width=0.9\textwidth]{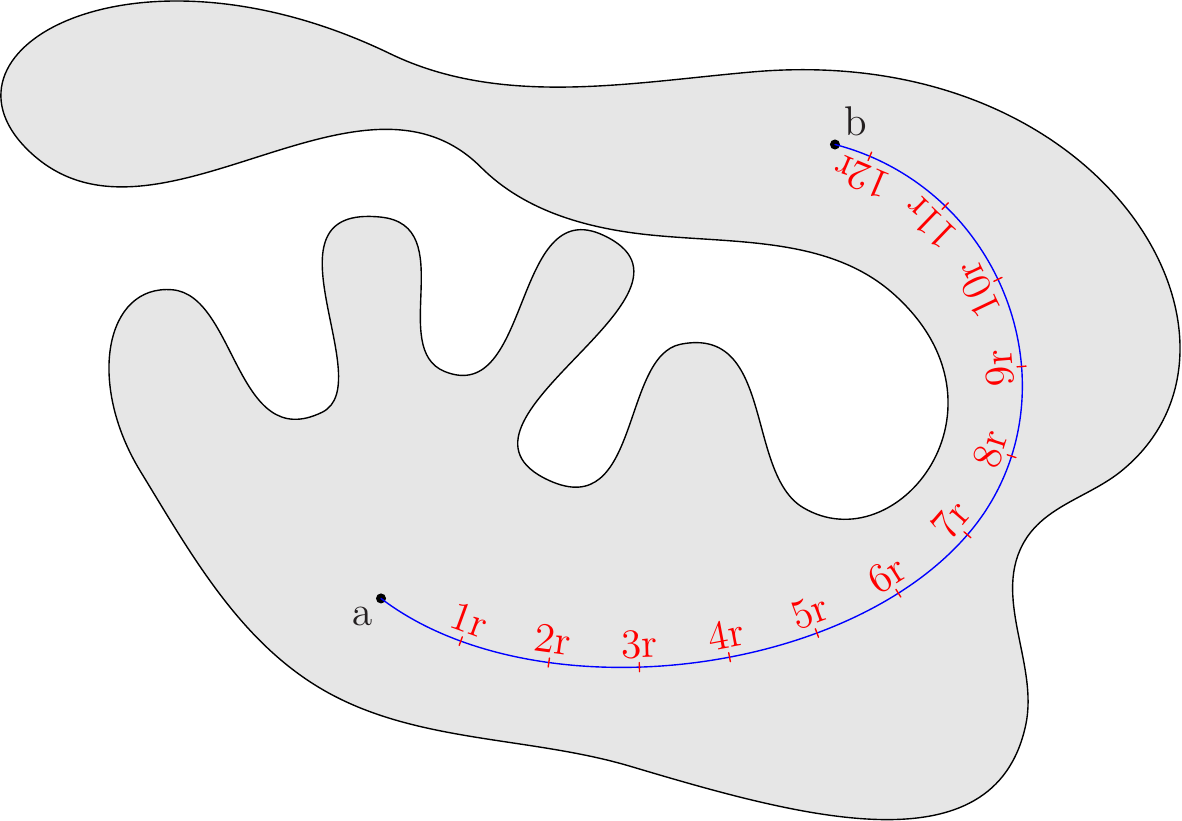}
  \caption{Difference between marker values of any two points is bounded.}
  \label{fig:proximity_global}
\end{figure}

\begin{Lemma} \label{lem:max_diff}
  The difference between marker values of any two points is bounded at
  any time instance.
  \[\forall t; \forall a,b \in\Omega\text{ : }|\sigma(a,t) - \sigma(b,t)| \leq \left\lceil \frac{d}{r} \right\rceil,\]
  where $d$ is the diameter of $\Omega$.
\end{Lemma}
\begin{Proof}
  Let us consider a path connecting the points $a$ and $b$. We can
  always split the path into sub-paths of length $r$, as depicted in
  Figure~\ref{fig:proximity_global}. According to
  Lemma~\ref{lem:close_points} difference between the marker values at
  the endpoints of every such sub-path is limited by 1, hence, we can
  conclude  that
  \begin{equation}
    \label{eq:43}
    |\sigma(a,t) - \sigma(b,t)| \leq \left\lceil
    \frac{l}{r} \right\rceil, 
  \end{equation}
  where $l$ represents the length of the path. Obviously, among all
  paths connecting $a$ and $b$ the shortest path will provide best
  upper bound on the difference between the pheromone levels at $a$
  and at $b$. Since the longest geodesic path in $\Omega$ is limited
  by $d$ we obtain the desired result:
  \begin{equation}
    \label{eq:32}
    |\sigma(a,t) - \sigma(b,t)| \leq \left\lceil \frac{d}{r} \right\rceil.
  \end{equation}
\end{Proof}

It is implied by Lemma~\ref{lem:max_diff} that the difference
between marker values is bounded. Our next step will be to show that 
the maximal marker value tends to infinity as $t$ goes to
infinity. First, we prove that marker values can only grow and never
decrease.
\begin{Lemma} \label{lem:non_decreasing}
  Pheromone level values at any point form a non-decreasing sequence. That is 
  \[ \forall t; \forall u\in\Omega\text{ : }\sigma(u,t+1) \geq
  \sigma(u,t). \]
\end{Lemma}
\begin{Proof}
  Let us assume the contrary, i.e., there exists a point $u\in\Omega$
  and time instance $t$  such that the pheromone level of $u$
  decreases during the $t$-th step:
  \begin{equation}
    \label{eq:decrease}
    \sigma(u,t+1) < \sigma(u,t).
  \end{equation}
  Let us now look at point $p_t$ -- the location of the robot at time
  $t$.  Obviously 
  $u$ belongs to $D(r,p_t)$ (otherwise it could not change its value), hence
  $\|u-p_t\|<r$. Assume that the minimal marker value among all
  points in $R(r,2r,p_t)$ was attained at some point $x$. We know also
  that $\sigma(p_t, t) \leq \sigma(x,t)$; otherwise, the robot does
  not change the pheromone values. Thus we have
  \begin{equation}
    \label{eq:decrease_2}
    \left \{
      \begin{array}{l}
        \sigma(x,t) + 1 = \sigma(u,t+1) < \sigma(u,t)\\
        \sigma(p_t,t) \leq \sigma(x,t)\\
        \|u - p_t\| < r
      \end{array}
    \right.
  \end{equation}
  This implies
  \begin{equation}
    \label{eq:decrease_3}
    \left \{
      \begin{array}{l}
        |\sigma(u,t) - \sigma(p_t,t)| > 1\\
        \|u - p_t\| < r 
      \end{array}
    \right.
  \end{equation}
  which contradicts Lemma~\ref{lem:close_points}. 
\end{Proof}

Next, we  show that the maximal value of the pheromone level grows
together with the step number $t$.
\begin{Lemma}
  \label{lem:max_growth}
  At any time instance $t$ maximal pheromone level in $\Omega$ is
  bounded from below by $t/n$ for some constant $n$.
\end{Lemma}
\begin{Proof}
  Imagine that the domain $\Omega$ is tessellated into $n$ cells
  so that the diameter of every such cell is less than $r$ (for
  convex cells we can, alternatively, require that the diameter of the
  circumscribing circle must be less than $r$). Let us
  examine the following expression:
  \begin{equation}
    \label{eq:min_sum}
    S_t = \sum_{i=1}^n m_t^i - \sigma(p_t,t), 
  \end{equation}
  where $m_t^i$ is the minimal marker value over the $i$th cell at
  time $t$ and $\sigma(p_t,t)$ is the marker value at the robot's
  location $p_t$ at time instance $t$. It was shown in 
  Lemma~\ref{lem:non_decreasing} that marker values of any point
  inside $\Omega$ form a non-decreasing sequence. Hence we claim
  that 
  \begin{equation}
    \label{eq:growth_sum}
    S_{t+1} > S_t.
  \end{equation}
  Indeed, for the non-marking step, the sum of the minima does not
  change: $\sum_{i=1}^n m_t^i = \sum_{i=1}^n m_{t+1}^i$; however
  $\sigma(p_t,t) > \sigma(p_{t+1}, t+1)$, and therefore, $S_{t+1} >
  S_t$. For the marking step, assuming
  that the robot goes from  cell $k$ into  cell $l$, we have
  $\sigma(p_t, t) \leq \sigma(p_{t+1}, t+1)$, and therefore,  $m_t^k
  \leq \sigma({p_{t+1}, t+1})$. Additionally, the whole cell $k$ was
  marked during this step and thus $m_{t+1}^k = \sigma(p_{t+1}, t+1)
  + 1$. Hence we have
  \begin{equation}
    \left \{
      \begin{array}{l}
        m_t^k - \sigma(p_t, t) \leq 0\\
        m_{t+1}^k - \sigma(p_{t+1}, t+1) = 1.
      \end{array}
    \right. 
  \end{equation}
  Since the sum of the other minima cannot decrease as was shown in
  Lemma~\ref{lem:non_decreasing}, we conclude again that $S_{t+1} >
  S_t$.
  Given that $S_0 = 0 $, we readily conclude that
  \begin{equation}
    \label{eq:bound_sum}
    S_t \geq t \Rightarrow \sum_{i=1}^n m_t^i \geq t \quad \forall t,
  \end{equation}
  which leads us to the conclusion that there exists $k \in
  1,2,\ldots, n$ such that
  \begin{equation}
    \label{eq:42}
    m^k_t \geq \frac{t}{n}.
  \end{equation}
  Hence the lemma is proved.
\end{Proof}
At this point we are ready to prove the main result of this work.
\begin{Theorem}  \label{theo:the_one}
  The domain $\Omega$ will be covered within a finite number of steps.
\end{Theorem}
\begin{Proof}
  According to Lemma~\ref{lem:max_growth} after
  $n\lceil\frac{d}{r}\rceil+1$ steps, at least 
  one of the $m^i$ values will be greater than
  $\lceil\frac{d}{r}\rceil$ and thus the whole domain will be covered.
\end{Proof}

\section{MAW - Efficiency Analysis}
\label{sec:maw-effic-analys}
As we proved earlier the domain $\Omega$ will be covered by a single
robot after $n\lceil\frac{d}{r}\rceil+1$ steps where $d$ is the
diameter of the domain, $r$ - the covering radius of the robot's effector
and $n$ - the number of cells in some tessellation of $\Omega$ (see
Lemma~\ref{lem:max_growth} for the definition of $n$).
We shall now analyze how good this upper bound is. In order to make such
comparison we shall find an approximation to  $n$ and to
find out what is the best upper bound possible.

\begin{figure}[H]
  \centering
  \includegraphics[width=0.4\textwidth]{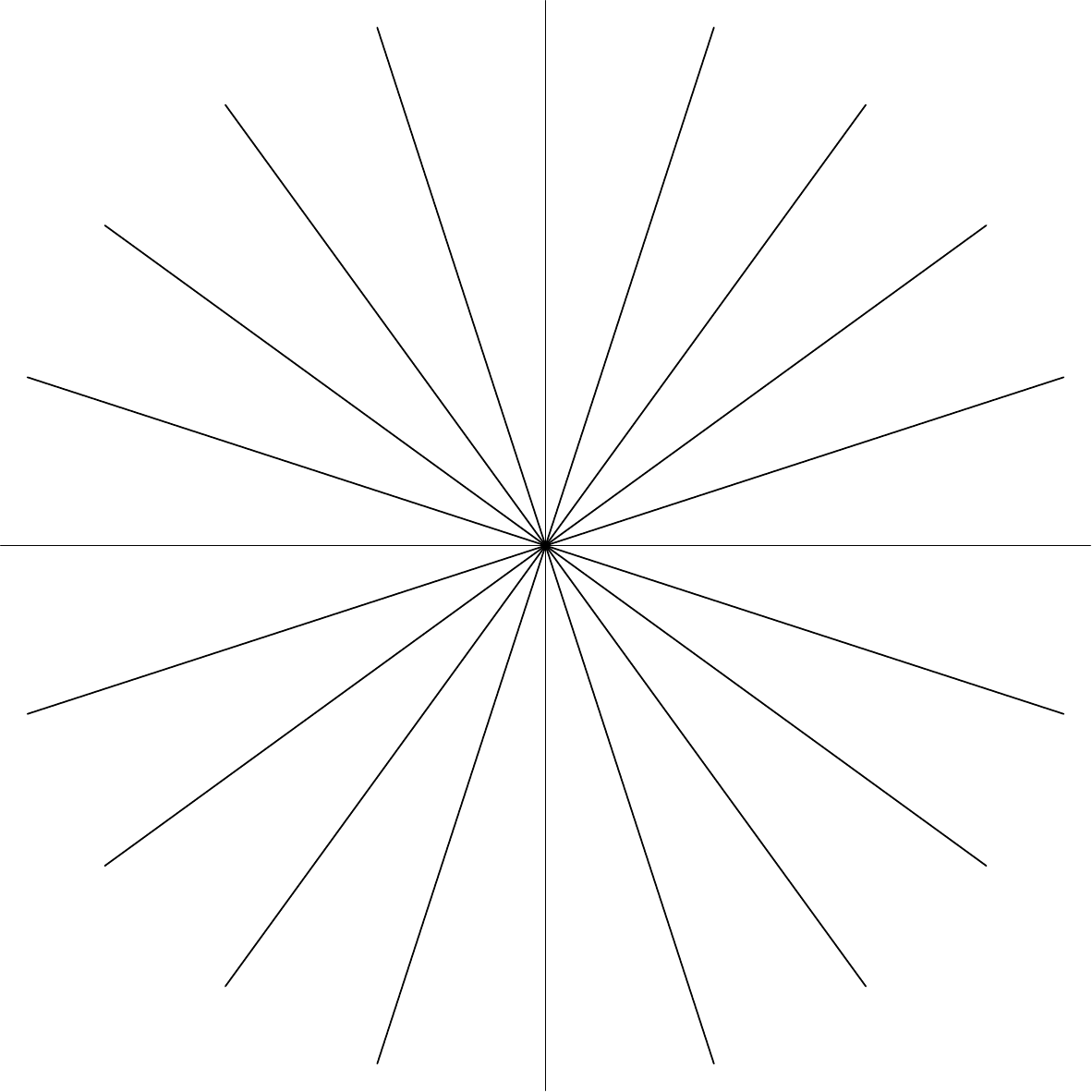}
  \caption{A ``pathological'' star shaped domain.}
  \label{fig:star_shaped}
\end{figure}

Let us denote by $N_{r}$ the minimal number of steps required by the robot
to cover the whole domain $\Omega$ (here we only assume that the robot
can cover an open disk of radius $r$ at every step, however, no
assumption is made regarding the algorithm governing the robot's
behavior). Clearly $ N_{r} = \frac{A_\Omega}{a_r}$, 
where $A_\Omega$ and $a_r$ are the areas of the domain $\Omega$ and
robot's effector respectively. Obviously, this is an ultimate lower
bound and no algorithm can beat it. However this bound fails be tight
enough for domains whose shape factor (ratio of the squared domain
perimeter to its area multiplied by $4\pi$) is far from 1 (not round). For
example, domains, comprised of finite number of curves and line
segments would have zero area providing lower bound to be zero, which
is, of course, meaningless. As a possible solution, in~\cite{wagner-mac}
the authors used the area of the ``augmented'' domain
$\bar{\Omega}$, which results from an inflation or expansion $\Omega$ in all
directions by $r$ , i.e., $\Omega$ has undergone a
morphological dilation with a disk of radius $r$. In this case we
have~\cite{wagner-mac}:
\begin{equation}
  \label{eq:omega_area}
  A_{\bar{\Omega}} \leq  A_{\Omega}+rP_{\Omega}+\pi r^2,
\end{equation}
where $A_{\Omega}$ and $P_{\Omega}$ are the area and the perimeter
of $\Omega$, respectively. This
approach has a serious drawback: there 
are situations when the augmented domain can be covered by the robot
in a finite number of steps while the original domain can
not. As an illustration of such domain consider a star-shaped domain
comprised of a number of line segments, say of length $r$,
emanating from a common origin, as shown in
Figure~\ref{fig:star_shaped}. Obviously, the number of line segments
can be infinite and since we cover an open dist of radius $r$ in each
step we will be forced to ``enter'' into every such line segment in
order to cover it completely and thus the number of step required to
cover such a domain will also be infinite. Note, however, that the
augmented domain in this case can be covered in a finite number of steps.

We suggest another way of performance assessment which is correct by
construction and does not depend on geometric properties like area or
perimeter. Consider best possible algorithm that covers in each step
an open disk of radius $r/2$, besides this requirement we do not limit the
algorithm in any other way. Assume that the our domain $\Omega$ can be
covered by such algorithm. That is there exists a finite sequence of points
$P_1, P_2, \ldots P_N$ of robot's locations that results in complete
coverage, that is :
\begin{equation}
  \label{eq:145}
  \Omega\subset \bigcup_{i=1}^{N} D
  \left(
    \frac{r}{2}, P_{i}
  \right).
\end{equation}
Alternatively, we can say that for every point $a$ in $\Omega$
there exist a number $k$ ($1\leq k \leq N$) such that $a \in D(r/2,
P_{k})$. Now, if we consider the best possible algorithm, as before we
denote its coverage time by $N_{r/2}$, we easily conclude that
the upper bound on the number of steps required by MAW algorithm is:
\begin{equation}
  \label{eq:26}
  N_{r/2}\left\lceil\frac{d}{r}\right\rceil+1.
\end{equation}
Indeed, if we consider a particular coverage path or the robot
described by the sequence of its successive positions: $P_{1}, P_{2},
\ldots P_{N_{r/2}}$ we always can perform Voronoi tessellation around these
points. Each cell in this tessellation will have diameter smaller than
$r$ and thus this tessellation will be like the one we used in
Lemma~\ref{lem:max_growth}.

We have only to estimate the $d/r$
fraction. Obviously, $\frac{d}{r} \leq N_{r/2}$ on one hand, on the
other hand we can 
estimate the lower bound: $\frac{d}{r} \geq
\frac{\pi}{4}\sqrt{N_{r/2}}$ (for domains
that have a shape close to a circle). Hence we have shown that the
upper bound time is polynomial with respect to best possible time of
algorithm whose covering radius is $r/2$.
\begin{equation}
  \label{eq:27}
  t_{\mathrm{coverage}} \leq N_{r/2}^x,
\end{equation}
where $0.5 \leq x \leq 2$.

The main question here is whether we can conclude that
\begin{equation}
  \label{eq:45}
  t_{\mathrm{coverage}} \leq N_{r}^x,
\end{equation}
i.e., whether our coverage time is bounded by a polynomial function of
$N_{r}$, which is best possible coverage time among all algorithms
whose covering radius is $r$. The general answer is ``No''. In fact, a
general theorem regarding limitations of such a type of algorithms:
\begin{Theorem}
  \label{theo:efficency}
  Given an algorithm whose marking area is an open disk of radius $r$
  and step size is grater than or equal to $r$. The time
  for complete coverage $t_{\mathrm{coverage}}$ of a continuous domain is (tightly) limited
  from below by $N_{r/2}$. Moreover $t_{\mathrm{coverage}}$ cannot be
  expressed as a bounded function of $N_{x}$ for any  $x > r/2$. 
\end{Theorem}
\begin{Proof}
  The proof is by example of such a domain: we consider again a domain
  comprised of $n$ line segments emanating common origin $O$, as shown if
  Figure~\ref{fig:star_shaped}. Assuming that the length of each line
  segment is $r/2$ we easily verify that $N_{r/2}= n $ and that
  $t_{\mathrm{coverage}} = n$ (assuming that the robot's initial
  position was an end-point of any line segment. Hence
  $t_{\mathrm{coverage}}$ is (tightly) bounded from below by
  $N_{r/2}$. Now if we assume that $n$ is infinite we easily conclude
  that such domain cannot be covered in a finite number of steps by
  our algorithm (again, provided that the initial robot location was
  at an end point of any line segment), while this domain can be
  covered in two steps by an optimal algorithm whose covering radius
  is greater than $r/2$ (we simply go to the origin $O$ in the very
  first step and the whole domain will be covered in the next step).
  Hence, the theorem is proved. 
\end{Proof}

Of course the above theorem is quite general, for a particular domain
one can have
\begin{equation}
  \label{eq:21}
  N_{r}\propto N_{r/2}, 
\end{equation}
hence, best possible time would be linear in terms of $N_{r}$ and
MAW's coverage time would be limited from above by $N_{r}^{2}$ times
some constant.

Let us elaborate more about the relationship between $N_{r}$ and
$N_{r/2}$. Consider a domain of area $A$ and perimeter $P$. Consider
also an optimal coverage with radius $r$. According to our definitions
this coverage requires exactly $N_{r}$ steps. Let us now look at the
Voronoi tessellation around corresponding robots' locations, there are
$N_r$  cells in this tessellation. Cells that do not include
the boundary of the domain are convex, those that do include domain's
boundaries may not be convex. Each convex cell in this tessellation
can be covered by a finite (and well-defined) number of disks of
radius $r/2$, hence for these cells we can conclude that $N_{r}
\propto N_{r/2}$. For non-convex cells this claim may not be correct,
thus, for domains whose tessellation consists mainly of convex cells
we have approximately $N_{r} \propto N_{r/2}$. Note that a similar analysis
was carried out  on the MAC algorithm~\cite{wagner-mac} where the
authors claim that MAC algorithm is asymptotically linear for domains
that obey $A>>Pr$. This is, fact, equivalent to say that the number of
cells in the Voronoi tessellation described above that do not
contain domain's boundary is large compared to the number of the cells
that do.

\chapter{Extensions}
\label{sec:extensions}
\section{Repetitive Coverage}
\label{sec:repetitive-coverage}
In some scenarios we might be interested in repetitive coverage of the
domain. For example, repetitive coverage is necessary in the
aforementioned scenario
when robots perform minefield de-mining and their mine detection is
not perfect, i.e., the probability of detecting a mine when  the robot's sensors
sweep above it is less than one. In this case repetitive coverage is
required to minimize the probability of leaving any mines 
undetected. In this case,  we have to
give an upper bound on time between two successive visits of the
robot in order to guarantee an improvement in detection
probability. This requirements also arises naturally in tasks such as 
surveillance or patrolling where robots are required to visit every
point over and over  and the time between two successive visits must
be limited by a constant. In other word we shall show now that our
algorithm has the property of patrolling.
Before we start proving this result we shall need the following lemma.
\begin{Lemma}
  \label{lem:sum_to_time}
  For any two time instances $t_1$ and $t_2$, if  $t_2 > t_1$ then
  the following inequality must hold:
  \[
  S_{t_2} - S_{t_1} \geq t_2 - t_1.
  \]
\end{Lemma}
\begin{Proof}
  Let us write $t_2 = t_1+n$ for some
  natural $n$ and prove the lemma by  mathematical induction on
  $n$. For  $n=1$ the lemma holds due to 
  Equation~(\ref{eq:growth_sum}). Assuming that the lemma holds for
  some $n$, we can easily conclude that the lemma holds for $n+1$ as
  well. 
\end{Proof}

\begin{figure}[H]
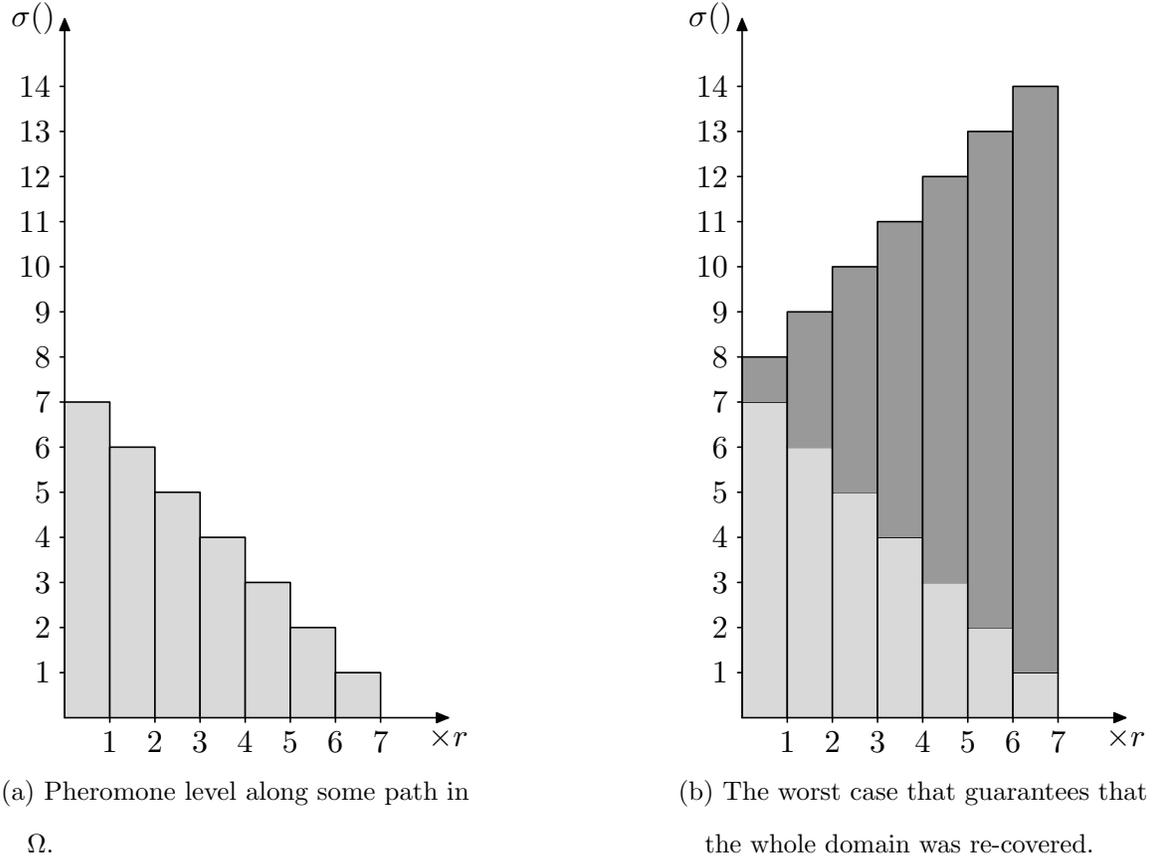

  \centering
  \subfloat[Pheromone level along some path in $\Omega$.]
  {\label{fig:recovered-a} \includegraphics[width=0.4\textwidth]{maw.6}}
  \hfill
  \subfloat[The worst case that guarantees that the whole domain was
  re-covered.]
  {\label{fig:recovered-b} \includegraphics[width=0.4\textwidth]{maw.7}}
  \caption{Repetitive coverage illustration}
  \label{fig:recovered}
\end{figure}

\begin{Theorem}
  \label{theo:repetitive}
  For any point $a \in \Omega$, the time period between two successive
  visits of the robot is bounded from above by
  \[2n
  \left(
    \left\lceil
      \frac{d}{r}
    \right\rceil
    +1
  \right)
  \].
\end{Theorem}
\begin{Proof}
 If we show that after a sufficient time period the pheromone level
 changes at all locations in the domain $\Omega$, we can obviously be
 sure that all points were re-visited by the robot during this time
 period. Let us look at time instance $t_s$ when the robot covers our
 point of interest $a$. We denote by $\sigma_{max}(t_s)$ the maximal
 pheromone level over $\Omega$ at that time. If we show that at some
 time instance  $t_e$ the minimal pheromone level denoted by
 $\sigma_{min}(t_e)$ becomes  greater than the maximal value that was
 at time $t_s$:
 \begin{equation}
   \label{eq:6}
   \sigma_{min}(t_e) > \sigma_{max}(t_s), 
 \end{equation}
 then we can easily conclude that during the time period $t_e - t_s$
 the pheromone level  changed at all points and thus all points
 (including $a$) were re-covered by the robot. Let us examine $S_{t_s}$
 and $S_{t_e}$ as defined in the Equation~(\ref{eq:min_sum}).
 On the one hand:
 \begin{equation}
   \label{eq:8}
   S_{t_s} = \sum_i^n m^i_{t_s} - \sigma(p_{t_i}, t_i)
   \geq \sum_i^n  m^i_{t_s}
   \geq \sum_i^n  \sigma_{min}(t_s)
   = n\,\sigma_{min}(t_s)
 \end{equation}
 According to  Lemma~\ref{lem:max_diff}
 \begin{equation}
   \label{eq:9}
   \sigma_{min}(t_s) \geq \sigma_{max}(t_s) +
   \left\lceil\frac{d}{r}\right\rceil.
 \end{equation}
 Combining Equations~(\ref{eq:8}) and (\ref{eq:9})  we get
 \begin{equation}
   \label{eq:10}
   S_{t_s} \geq n
   \left(
     \sigma_{max}(t_s) -
     \left\lceil
       \frac{d}{r}
     \right\rceil
   \right).
 \end{equation}
 On the other hand we want to know the time instance $t_e$ that
 guarantees that $\sigma_{min}(t_e) \geq \sigma_{max}(t_s)+1$. Instead
 of estimating $t_e$ directly from $\sigma_{min}(t_e)$,  we shall
 look for $t_e$ that guarantees the existence of $\sigma$ value greater
 than or equal to $\sigma_{max}(t_s) + 1 + \lceil\frac{d}{r}+1\rceil$,
 which guarantees by Lemma~\ref{lem:max_diff} that 
 $\sigma_{min}(t_e) \geq \sigma_{max}(t_s)+1$. Now, in the same
 way as the proof of Theorem~\ref{theo:the_one}, we can say
 that once $S_{t_e}\geq n(\sigma_{max}(t_s) + 1 + \lceil\frac{d}{r}\rceil+1)$,
 we have $\sigma_{min}(t_e)\geq \sigma_{max}(t_s) + 1$. Thus we
 have
 \begin{equation}
   \label{eq:11}
   S_{t_e} - S_{t_s} \leq
   n
   \left(
     \sigma_{max}(t_s) + 1 +
     \left\lceil
       \frac{d}{r}
     \right\rceil
     +1
   \right)
   -
   \left(
     \sigma_{max}(t_s) -
     \left\lceil
       \frac{d}{r}
     \right\rceil
   \right)\\
   = 2n
   \left(
     \left\lceil
       \frac{d}{r}
     \right\rceil
     +1
   \right).
 \end{equation}
 According to Lemma~\ref{lem:sum_to_time} we have
 \begin{equation}
   \label{eq:12}
   t_e - t_s \leq S_e - S_s \leq
   2n
   \left(
     \left\lceil
       \frac{d}{r}
     \right\rceil
     +1
   \right),
 \end{equation}
 which completes the proof.
\end{Proof}
We now make two important observations. First, the upper time
bound between two successive visits by the robot does not depend on
the current pheromone level distribution. This is  determined completely by
the geometric parameters of the problem: $r$, $n$, and $d$. Second, we
observe that this time limit is twice as long as the time period
needed for complete coverage. This situation is quite
intuitive. Indeed, observe the pheromone level along some path
in $\Omega$, such as the one shown in Figure~\ref{fig:recovered-a}. In
this case the robot may start by ``filling'' the hollow area on the
right until it becomes a hill and  only then covers the left-hand part,
which used to be a summit point and has now became the lowest point in
the pheromone level profile as shown in Figure~\ref{fig:recovered-b}.
\section{Noise Immunity}
\label{sec:noise-immunity}
\begin{figure}[H]
  \centering
  \includegraphics[width=0.4\textwidth]{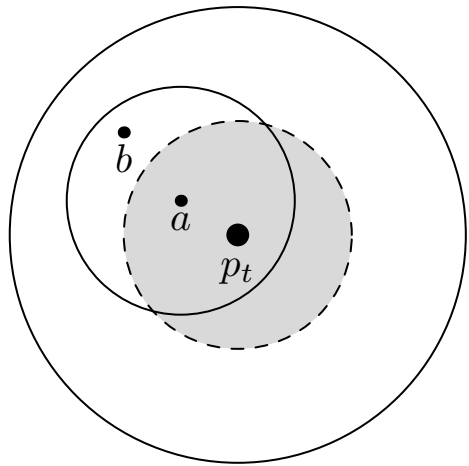}
  \caption{Close points in noisy environment}
  \label{fig:noise_proximity1}
\end{figure}

So far we always assumed that there is no noise in the input,
i.e., the robot starts with a domain that does not contain any pheromone
marks. Unfortunately, in the real life such a clean environment is not
always available. For example, spurious pheromone marks may arise as a
result of previous attempts to explore the domain by similar robots,
which might have used other algorithms and thus the initial pheromone level
distribution does not necessarily obey the proximity principle. In
general we assume that the initial pheromone level distribution is
given by some function
\begin{equation}
  \label{eq:13}
  N:\Omega \mapsto \mathbb{Z}^*,
\end{equation}
where $\mathbb{Z}^*$ denotes the set of non-negative integers. As a short
digression we have to note that such initial pheromone marks pose a
severe problem to all trail-based algorithms. The reason is that
such algorithms, for the sake of efficiency, do not get close to their
own trails and thus any initial pheromone marks would be interpreted
as trails, resulting in uncovered areas around such marks. The result
may be even worse if such false trails split the domain into several
disconnected parts, in this case the robot will not be able to exit the
part where it was located initially. Our algorithm, on the contrary,
can easily overcome this problem as we prove below. Actual covering
times in presence of noise are demonstrated in
Section~\ref{sec:maw-noisy-envir}. Let us start with several
lemmas.
\begin{Lemma}
  \label{lem:noise-immunity-1}
  Immediately after  point $a\in\Omega$ has been  marked by the robot
  for any point $b\in\Omega$ such that $\|a - b\| \leq r$ we have:
  \begin{equation}
    \label{eq:7}
    \sigma(b,t+1) \geq \sigma(a,t+1) - 1,
  \end{equation}
  where $t$ denotes the time instance when the new pheromone level was
  assigned to $a$. 
\end{Lemma}
\begin{Proof}
  Since the pheromone value of $a$ changes during the $t$th step we
  conclude that $a\in D(r, p_t)$, where $p_t$ denotes robot's location
  at time $t$. We know also that $\|a - b\|\leq r$ and thus $b$
  either belongs to $D(r, p_t)$ or to $R(r, 2r, p_t)$ (see
  Figure~\ref{fig:noise_proximity1}). Hence there are two possible
  scenarios: either $b$ belongs to $D(r, p_t)$ or $b$ belongs to $R(r,
  2r, p_t)$. In the former case $\sigma(b,t+1) = \sigma(a,t+1)$ since the
  algorithm assigns the same value to all points in $D(r, p_t)$ and
  the lemma clearly holds. In the latter case ($b\in R(r, 2r, p_t)$)
  we recall that the algorithm seeks for the minimal pheromone value
  inside $R(r,2r,p_t)$, say attained at some point $x$ and set new
  pheromone level inside $D(r,p_t)$ to be equal to $\sigma(x,t) +
  1$. Hence, we have:
  \begin{equation}
    \label{eq:17}
    \left\{
      \begin{array}{l}
        \sigma(b,t+1) = \sigma(b,t) \geq \sigma(x,t)\\
        \sigma(a,t+1) = \sigma(x,t) + 1
      \end{array}
    \right. .
  \end{equation}
  This leads us again to the conclusion
  \begin{equation}
    \label{eq:18}
    \sigma(b,t+1) \geq \sigma(a,t+1) - 1 . 
  \end{equation}
  Hence the lemma is proved.
\end{Proof}
Though this lemma resembles Lemma~\ref{lem:close_points}, it does
not guarantee that the proximity principle is obeyed in noisy
environment. We demonstrate a stronger result later in
Lemma~\ref{lem:noise-immunity-2}.

At this moment we shall prove that the pheromone level at marked
points, i.e., pheromone left by the robot never decreases.
\begin{Lemma}
  \label{lem:noise-immunity-3}
  Pheromone level values at any marked point form a non-decreasing
  sequence; that is
  \begin{displaymath}
    \forall t\; \forall u \in\Omega : \sigma(u,t+1) \geq\sigma(u,t)
  \end{displaymath}
  given that $u$ was marked by the robot at the time prior to $t$.
\end{Lemma}
\begin{Proof}
  Let us assume the contrary, i.e., for some time instance $t$ and for
  some point $u\in\Omega$ we have:
  \begin{equation}
    \label{eq:22}
    \sigma(u,t) < \sigma(u, t+1)
  \end{equation}
  Let $t$ be the first such  time instance. As usual we denote by
  $p_t$ the robot's  location at time $t$. According to the MAW
  algorithm the robot seeks to the minimal pheromone level in $R(r,2r,
  p_t)$. Say this minimal level is attained at some point $x\in
  R(r,2r, p_t)$. Since $u$ changes its value during the $t$th step we
  conclude that
  \begin{equation}
    \label{eq:23}
    \sigma(p_t, t) \leq \sigma(x,t).
  \end{equation}
  Otherwise no change happens, according to the MAW algorithm. Sine
  all points in $D(r,p_t)$ get the same value in the marking step we
  conclude:
  \begin{equation}
    \label{eq:24}
    \sigma(u,t+1) = \sigma(p_t,t+1)=\sigma(x,t) +1 \geq
    \sigma(p_t,t)+1
  \end{equation}
  Moreover, according to our assumption:
  \begin{equation}
    \label{eq:25}
    \sigma(u,t) > \sigma(u,t+1)\Rightarrow \sigma(u,t) \geq\sigma(p_t)+2.
  \end{equation}
  Thus, if we assume that the pheromone level at some point $u$
  decreases at time instance $t$ Equation~(\ref{eq:25}) must
  hold. Showing that this inequality is wrong we actually get a
  contradiction to the assumption and thus prove the lemma. Let us
  look at the time instance $t_u$ when the current pheromone level of
  $u$ ($\sigma(u,t)$) was set. According to
  Lemma~\ref{lem:noise-immunity-1}:
  \begin{equation}
    \label{eq:28}
    \sigma(p_t,t_u+1) \geq \sigma(u,t_u+1) - 1.
  \end{equation}
  Since $t_u < t$ and $t$ was chosen to be the first time when the
  pheromone level at any  point in $\Omega$ decreases we conclude that
  \begin{equation}
    \label{eq:29}
    \sigma(p_t, t)\geq\sigma(p_t,t_u+1).
  \end{equation}
  Substituting it into Equation~(\ref{eq:28}) we get
  \begin{equation}
    \label{eq:30}
    \sigma(p_t, t) \geq \sigma(u,t) - 1. 
  \end{equation}
  And thus, the inequality in Equation~(\ref{eq:25}) does not hold. This
  contradiction completes the proof of the lemma.
\end{Proof}
\begin{Lemma}
  \label{lem:noise-immunity-2}
  If, at some time instance $t$, both $a$ and $b$ had been
  marked by the robot their pheromone levels obey the proximity
  principle,i.e.,
  \begin{equation}
    \label{eq:19}
    \text{if }|a-b|\leq r, \text{ then }|\sigma(a,t) - \sigma(b,t)| \leq 1
  \end{equation}
\end{Lemma}
\begin{Proof}
  Since both $a$ and $b$ had been marked prior to time instance $t$
  there are exist time instances $t_a$ and $t_b$ when $a$ and $b$ got
  their current pheromone levels accordingly. Applying
  Lemma~\ref{lem:noise-immunity-1} we get the following two equations:
  \begin{equation}
    \label{eq:31}
    \left\{
      \begin{array}{l}
        \sigma(b, t_a+1) \geq \sigma(a, t_a+1) - 1\\
        \sigma(a,t_b+1) \geq \sigma(b, t_b+1) - 1.
      \end{array}
    \right.
  \end{equation}
  Or, substituting $\sigma(b, t_b+1) = \sigma(b,t)$ and
  $\sigma(a,t_a+1) = \sigma(a,t)$
  \begin{equation}
    \label{eq:33}
    \left\{
      \begin{array}{l}
        \sigma(b,t_a+1) \geq \sigma(a, t) - 1\\
        \sigma(a,t_b+1) \geq \sigma(b, t) - 1.
      \end{array}
    \right.
  \end{equation}
  Since $t > t_a$ and $t >t_b$ we can apply
  Lemma~\ref{lem:noise-immunity-3}:
  \begin{equation}
    \label{eq:34}
     \left\{
      \begin{array}{l}
        \sigma(b,t) \geq \sigma(b,t_a+1) \\
        \sigma(a,t) \geq \sigma(a, t_b+1).
      \end{array}
    \right.
  \end{equation}
  Substituting it into Equation~\ref{eq:33} we get
  \begin{equation}
    \label{eq:35}
    \left\{
      \begin{array}{l}
        \sigma(b,t) \geq \sigma(a, t) - 1\\
        \sigma(a,t) \geq \sigma(b, t) - 1.
      \end{array}
    \right.
  \end{equation}
  Which means
  \begin{equation}
    \label{eq:36}
    |\sigma(a,t) - \sigma(b,t)| \leq 1
  \end{equation}
  Hence the lemma is proved.
\end{Proof}
\begin{Lemma}
  The maximal pheromone level tends to infinity as $t$ goes to infinity.
\end{Lemma}
\begin{Proof}
  \label{lem:maximal2}
  The proof is identical to the one in Lemma~\ref{lem:max_growth}. We
  again introduce a virtual tessellation of domain $\Omega$ into $n$
  cells so that every such cell can be inscribed into a circle of
  diameter less than $r$. And, as before, we look at the sum:
  \begin{equation}
    \label{eq:37}
    S_t = \sum_{i=1}^n m_t^i - \sigma(p_t,t)
  \end{equation}
  The only difference that this time $m_t^i$ denotes the minimal
  marker value that was set by the robot and not as a result of the
  noise. As in Lemma~\ref{theo:the_one} we get
  \begin{equation}
    \label{eq:38}
    S_t \geq t\ \forall t.
  \end{equation}
  Thus the lemma is proved.
\end{Proof}
\begin{Theorem}
  For any initial noise profile $N:\Omega\mapsto \mathbb{Z}^*$, the
  domain $\Omega$ will be covered after $n(M_N - m_N
  +\lceil\frac{n}{r}\rceil) +1$ steps, where $M_N$ and $m_N$ denotes
  the maximal and the minimal pheromone levels at time $t=0$,
  respectively. 
\end{Theorem}
\begin{Proof}
  Let us denote by $M_N$ and by $m_n$ respectively, the maximal and
  the minimal values of the initial pheromone level given by
  $N$. According to Lemma~\ref{lem:maximal2} the maximal pheromone
  level in $\Omega$ grows and will eventually reach the value of
  $M_N + \lceil\frac{n}{r}\rceil +1$. We claim that at this moment the
  whole domain is covered by the robot. Indeed, let us look at some
  point $a \in\Omega$ that got at step $t$ this pheromone
  level. According to Lemma~\ref{lem:noise-immunity-1} for any point
  $b\in\Omega$ such  that $\|a - b\| \leq r$ we have
  \begin{equation}
    \label{eq:39}
    \sigma(b,t+1) \geq \sigma(a,t+1)-1 = M_N +
    \left\lceil
      \frac{n}{r}
    \right\rceil.
  \end{equation}
  Since $M_N + \lceil\frac{n}{r}\rceil > M_N$ we conclude that all
  such points are covered. In the same manner we get that all points
  whose distance from $a$ is less than or equal to $2r$ are also
  covered. And so on, maximal distance between points in$\Omega$ is
  bounded by $d$, thus one any point reaches pheromone level of $M_N +
  \lceil\frac{n}{r}\rceil +1$ we assure that the minimal possible
  pheromone level is $M_N+1$ which means that the whole domain
  $\Omega$ has been covered. The time needed to cover the domain is
  \begin{equation}
    \label{eq:40}
    t_{cover} = n
    \left(
      M_N - m_N +\left\lceil\frac{n}{r}\right\rceil\right) +1,
   ,
  \end{equation}
  where $m_N$ denotes minimal pheromone level at time $t=0$.
  Since the whole domain is covered we can guarantee, by
  Lemma~\ref{lem:noise-immunity-2} the proximity principle is obeyed
  by any two points in $\Omega$ and further repetitive coverage is
  governed by Theorem~\ref{theo:repetitive}. 
\end{Proof}
\section{Multiple Robots}
\label{sec:multiple-robots}
As a natural extension we would like to analyze how the MAW algorithm
can be applied to multi-robot environments. First of all, we must
address problems such as collisions both between the robots themselves
(if we deal with physical robots and not programs) and between
different pheromone levels when two (or more) robots try to mark the
same point in the domain.

At the moment we shall assume that the clock phases of all robots are
slightly different so that no two robots are active at the same
time. Thus each robot sees other robots as regular stationary
obstacles and acts accordingly. With this approach we also have no 
problem of simultaneous attempts to set (probably different) pheromone
levels at particular location by multiple robots, since only one robot
is active at any given time.

Let us find the upper bound for complete coverage provided we have $n$
robots. Using the same notation as in  Equation~(\ref{eq:min_sum})
we have:
\begin{equation}
  \label{eq:14}
  S_t = \sum_{i=1}^n m_t^i - \sum_{j=1}^n\sigma(p_t^j,t), 
\end{equation}
where $p_t^j$ denotes the location of the $j$-th robot at time
$t$. Using exactly the same reasoning as before, we again obtain:
\begin{equation}
  \label{eq:15}
  S_{t+1} > S_t,
\end{equation}
and consequently
\begin{equation}
  \label{eq:16}
  S_t \geq t,
\end{equation}
which leads us to the same upper bound we got for a single robot. Hence
adding more robots does not necessarily guarantees better coverage
time. However  simulations (see Section~\ref{sec:simul-exper})
demonstrate that there is a substantial improvement when we use more
robots.  
\section{Generalization for  Other  Metrics}
\label{sec:non-eucl-metr}
\begin{figure}[H]
  \centering
  \subfloat[Measuring distance in the $xy$-plane]
  {\label{fig:metric_example1}\includegraphics[width=0.6\textwidth]{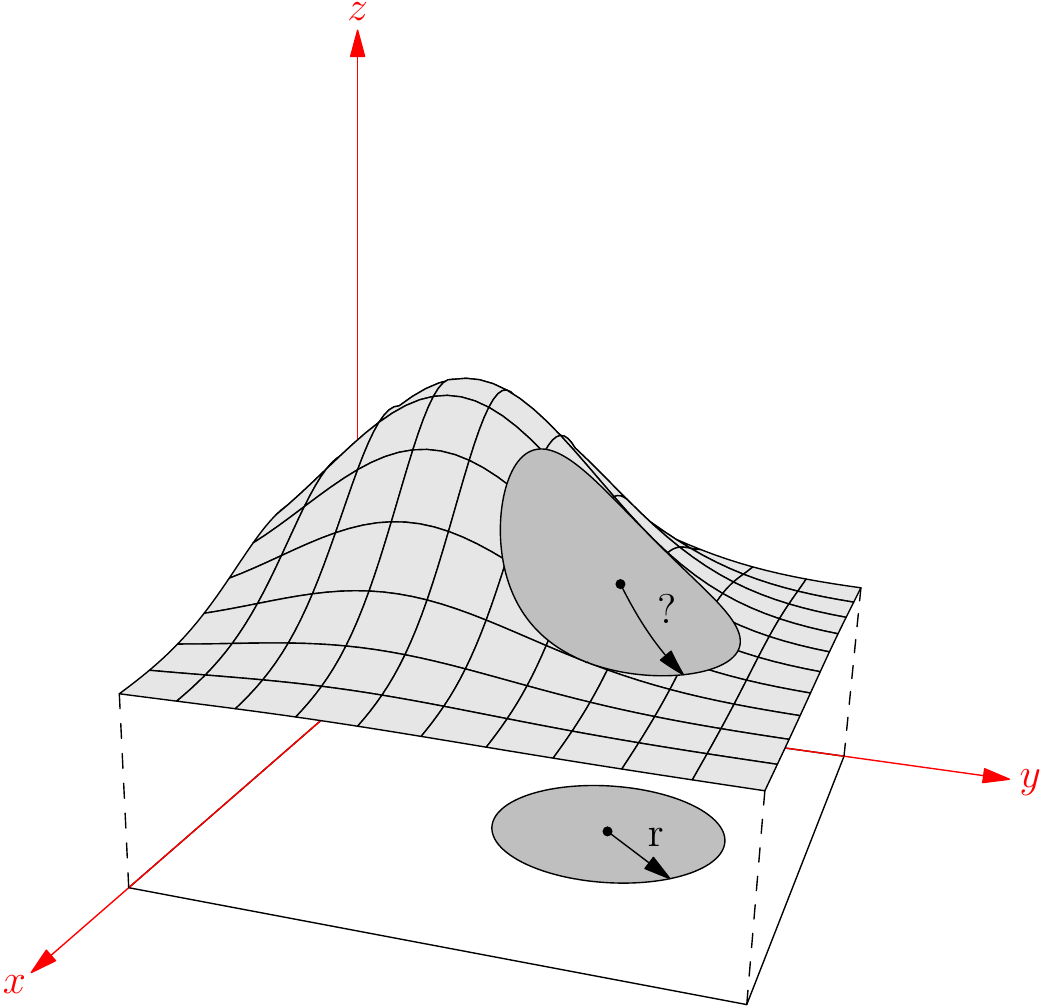}}\\
  \subfloat[Measuring distance on the surface.]
  {\label{fig:metric_example2}\includegraphics[width=0.6\textwidth]{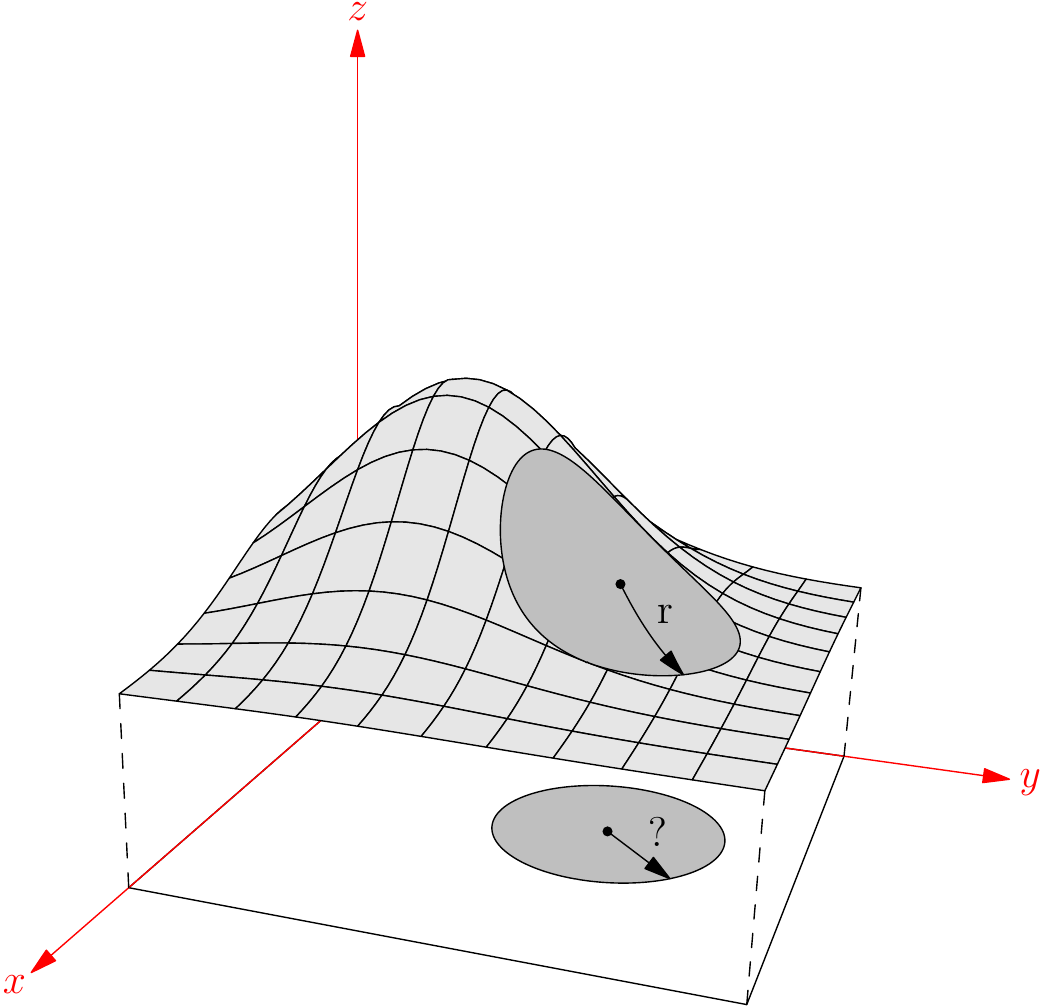}}
  \caption{Distance measurement.}
  \label{fig:metric_example}
\end{figure}

\begin{figure}[H]
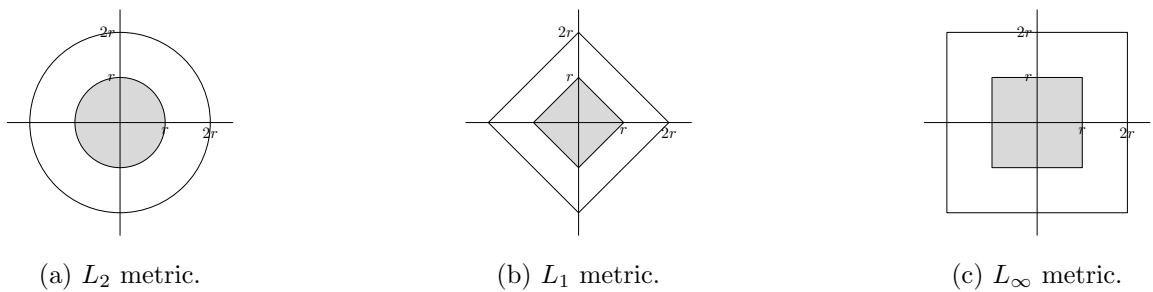

  \centering
  \subfloat[$L_2$ metric.]
  {\label{fig:metrics_l2}\includegraphics[width=0.2\textwidth]{maw.8}}
  \hfill
  \subfloat[$L_1$ metric.]
  {\label{fig:metrics_l1}\includegraphics[width=0.2\textwidth]{maw.9}}
  \hfill
  \subfloat[$L_\infty$ metric.]
  {\label{fig:metrics_linf}\includegraphics[width=0.2\textwidth]{maw.10}}
  \caption{Use of different metrics.}
  \label{fig:metrics}
\end{figure}

Until now we always assumed that the domain $\Omega$ is flat
two-dimensional domain, and  the usual Euclidean notion of the
distance was used. Nevertheless, we can consider a more complex case
of non-flat domains that looks like a surface  embedded in
$\mathbb{R}^3$ that can be described 
(at least locally) as a bi-variate function $z = f(x,y)$. In this case
we have a one-to-one correspondence between points in the $xy$-plane
and the points in the  $\mathbb{R}^3$ that lie on the surface. Hence
we can continue to measure the distance in the $xy$-plane (see
Figure~\ref{fig:metric_example1}) and once the corresponding domain is
covered in that plane the actual domain $\Omega$ will be covered
too. Or,alternatively we can measure the distance on the surface
itself (see Figure~\ref{fig:metric_example2}) in this case we have the
reverse situation: once the surface is 
covered the corresponding domain in the $xy$-plane will be covered as
well.
This simple example leads us to a more general conclusion: for
distance measurement we can use any  \emph{metric} $g$.
It is easy to verify that all the proofs remain valid
if we change the Euclidean, often referred to as $L_2$ distance to
another one. For example, we could use $L_1$ distance or,
alternatively, the $L_\infty$ distance which is particularly suitable
for computer simulations. Of course each choice of the metric changes
the form of the robot's effector. Three different forms, shown if
Figures \ref{fig:metrics_l2}, \ref{fig:metrics_l1}, and
\ref{fig:metrics_linf} correspond to $L_1$, $L_2$, and $L_\infty$
metrics accordingly.
Moreover we are not limited to 2-Dimensional spaces as the results
remain valid for higher dimensions, e.g., we can use the same
algorithm for covering 3D volumes assuming the robot's effector is a
ball of radius $r$ or, probably, a regular octahedron or a cube if we
choose to work with $L_1$ or $L_\infty$ metrics respectively.
\chapter{Simulations and Experiments}
\label{sec:simul-exper}
\section{General Notes}
\label{sec:sim-general-notes}
\begin{figure}[H]
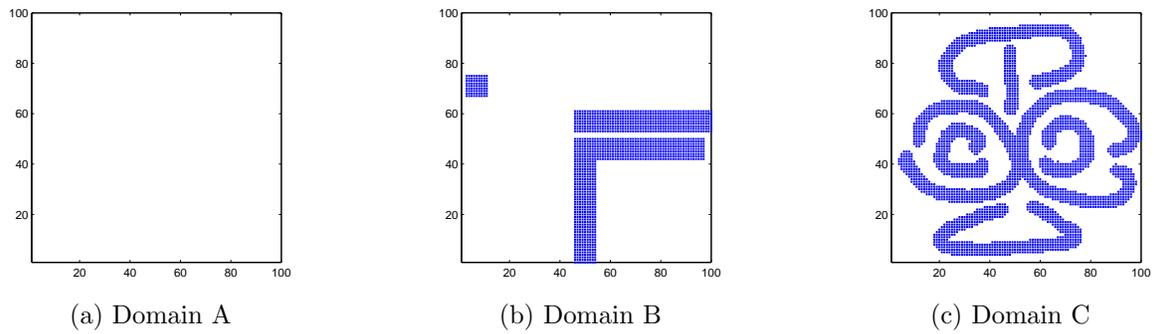

  \centering
  \subfloat[Domain A]%
  {\includegraphics[width=0.25\textwidth]{domain_a}}
  \hfill
  \subfloat[Domain B]%
  {\includegraphics[width=0.25\textwidth]{domain_b}}
  \hfill
  \subfloat[Domain C]%
  {\includegraphics[width=0.25\textwidth]{domain_c}}
  \caption{Simulation domains.}
  \label{fig:domains}
\end{figure}

\begin{figure}[H]%
  \centering
  \subfloat[Coverage: step 1]%
  {\includegraphics[width=0.3\textwidth]{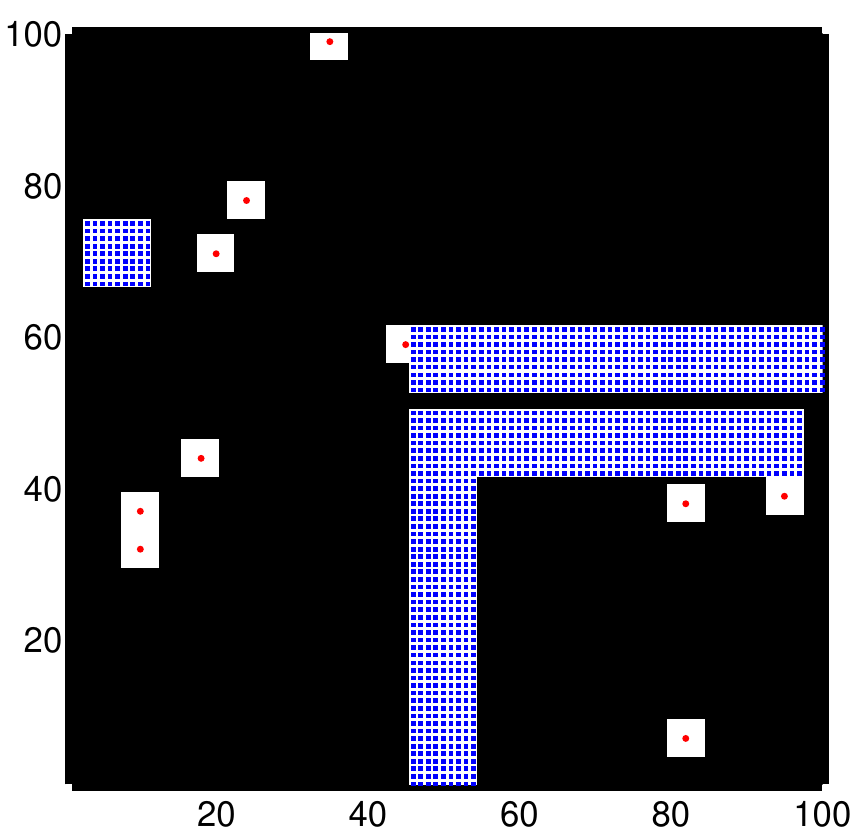}}%
  \hfill
  \subfloat[Odor map: step 1]%
  {\includegraphics[width=0.4\textwidth]{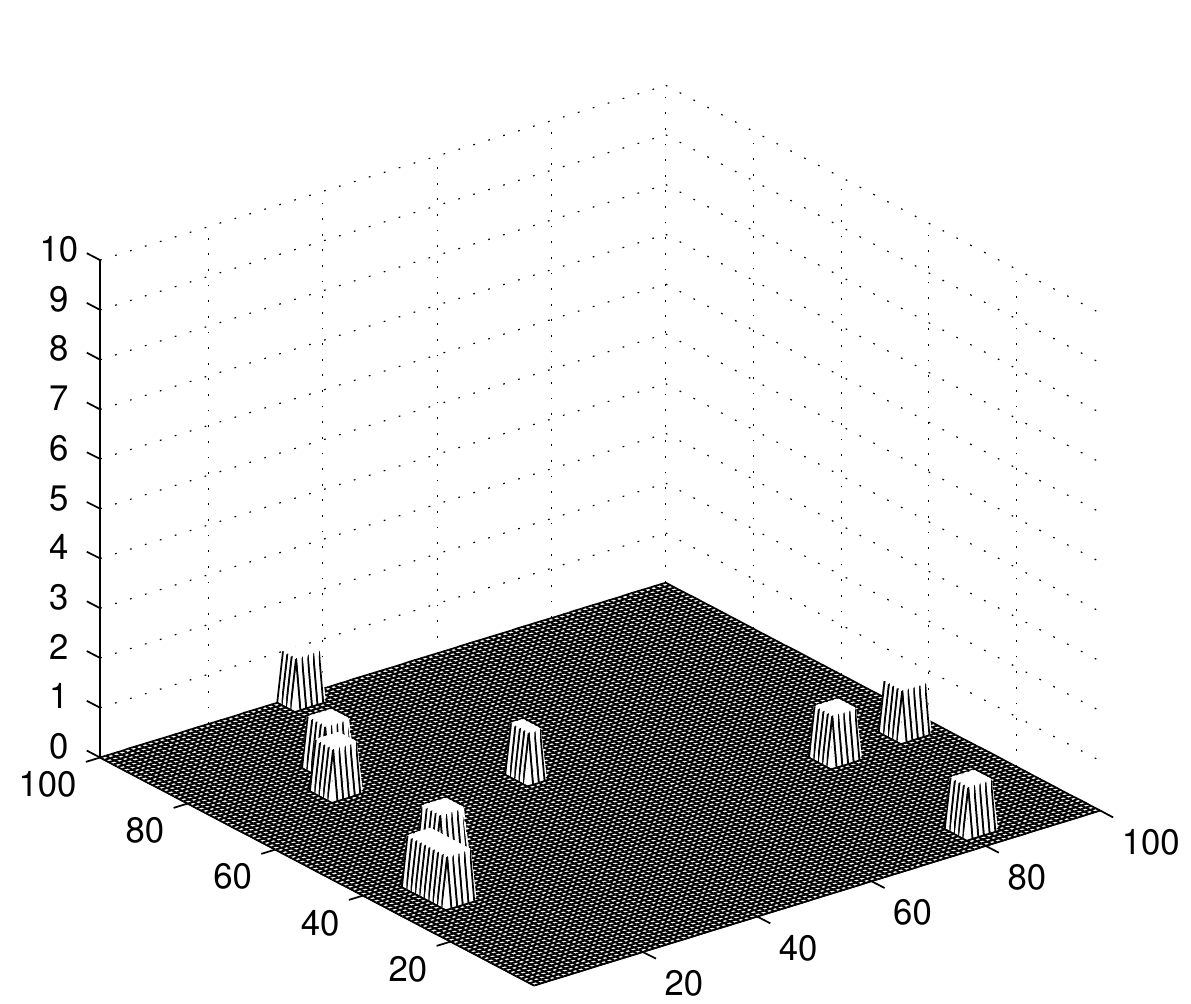}}\\
  \subfloat[Coverage: step  50]%
  {\includegraphics[width=0.3\textwidth]{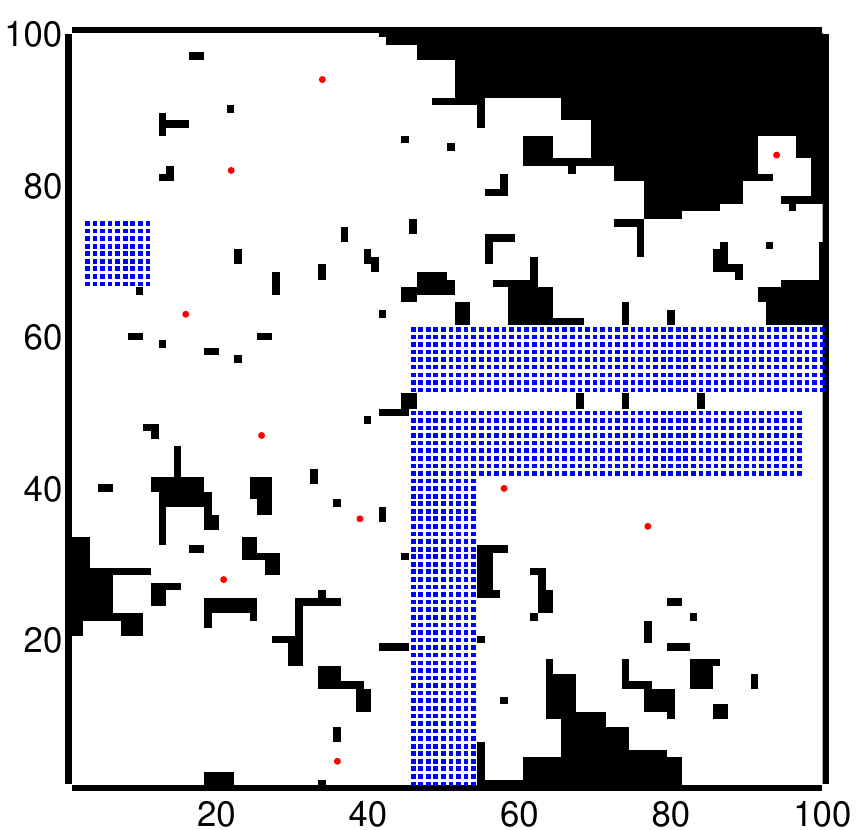}}%
  \hfill
  \subfloat[Odor map: step 50]%
  {\includegraphics[width=0.4\textwidth]{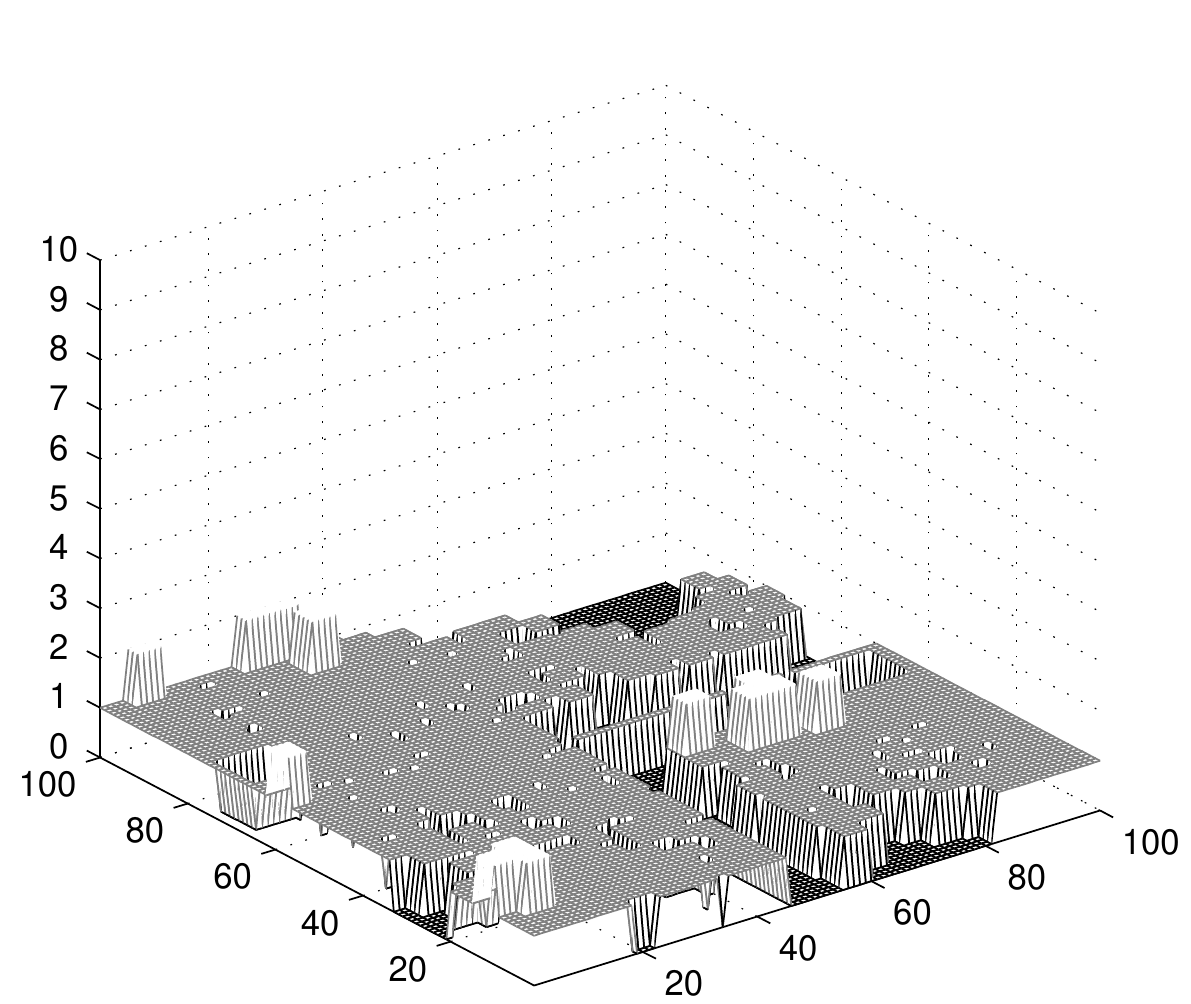}}\\
%  \caption{MAW progress on Domain B}
%  \label{fig:maw-progress}
% \end{figure}
% \begin{figure}[H]%
%   \centering
  \subfloat[Coverage: step 125]
  {\includegraphics[width=0.3\textwidth]{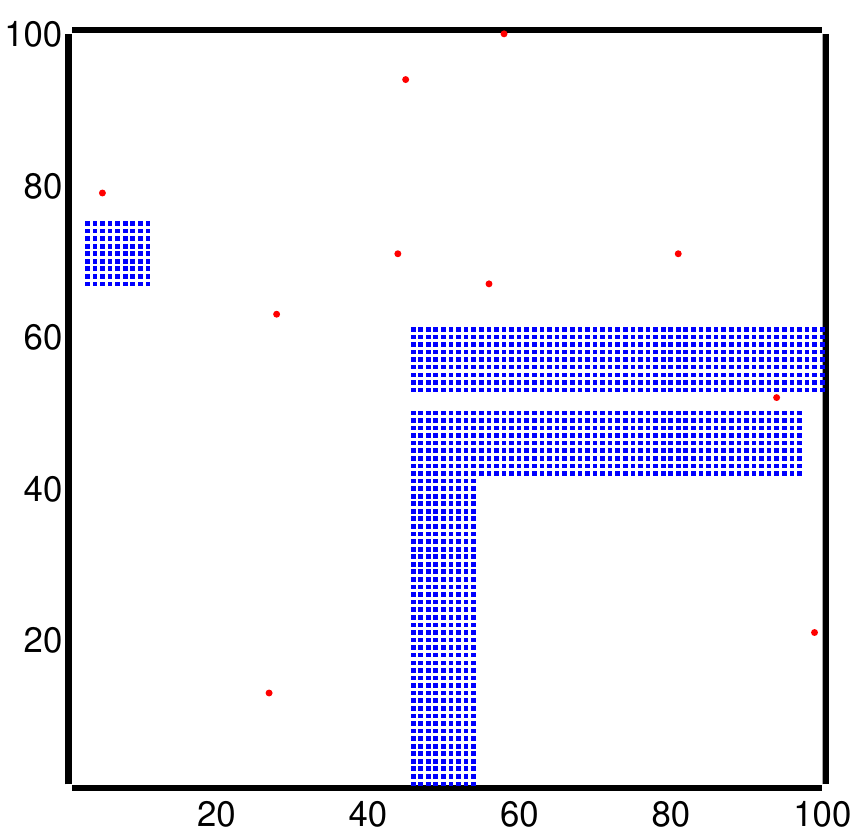}}%
  \hfill
  \subfloat[Odor map: step 125]
  {\includegraphics[width=0.4\textwidth]{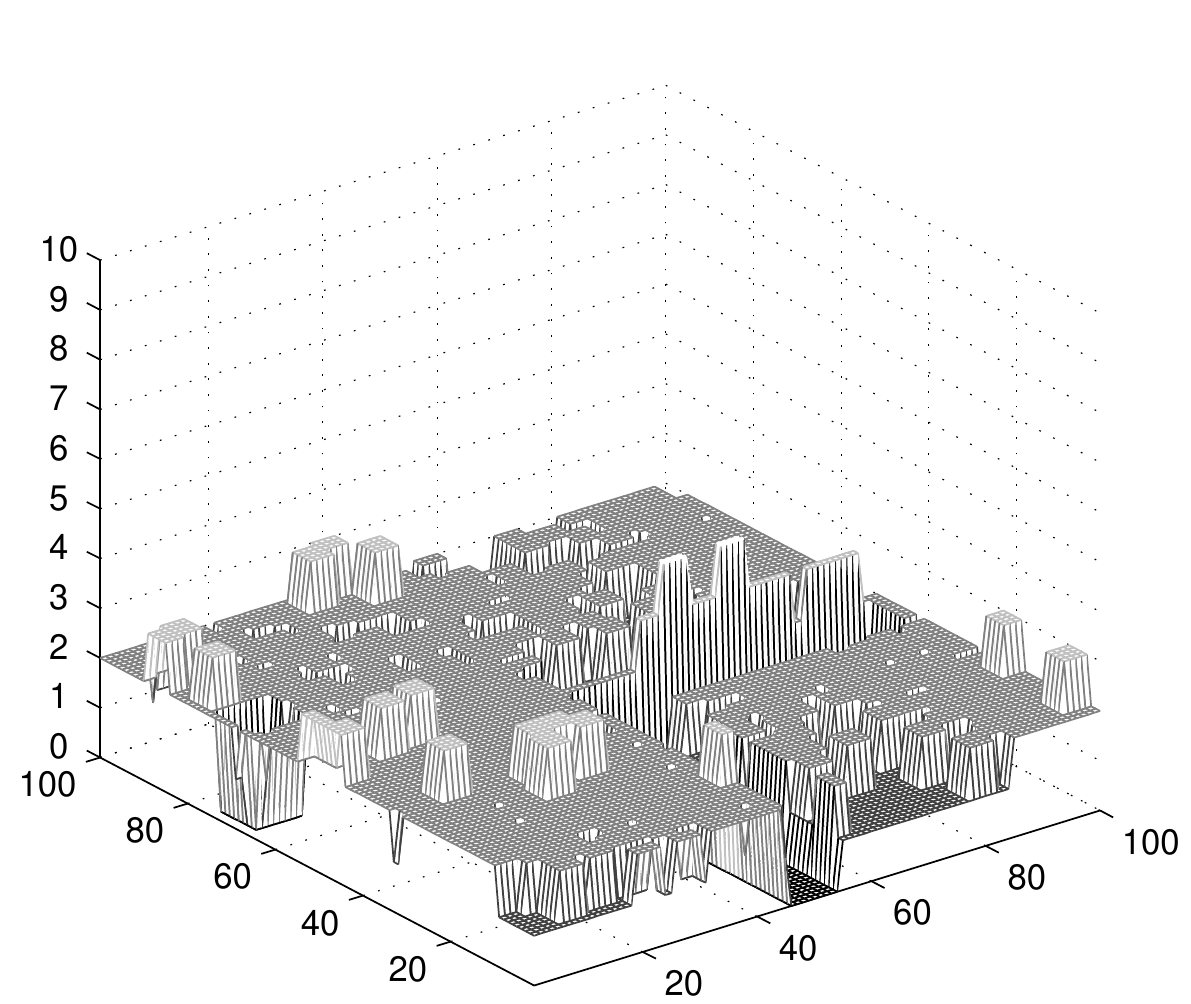}}
  \caption[]{MAW progress on Domain B}
  \label{fig:maw-progress}
\end{figure}

We used $L_\infty$ metric in our experiments
because the square shape of the effector and the sensing area that
correspond to this metric is particularly suitable for computer
simulations. In experiments with noise the robots were forced to start
at non-noisy location, i.e., at locations with minimal pheromone level
at time $t=0$. Additionally, in all experiments the robots were
modeled as points 
and multiple robots were allowed to occupy the same location. We
always measured the number of time steps until the robots
covered the domain for the first time, averaged over 100
runs. Experiments were conducted on
the domains shown in
Figure~\ref{fig:domains}.
All domains are of size $100\times 100$ pixels and marking radius in
all experiments was set to 3, i.e., each step robot marks a square of
$5\times5$ pixels. Figure~\ref{fig:maw-progress} demonstrates some
stages of covering Domain B by ten robots. 

\section{Comparing MAW to other algorithms}
\label{sec:comparing-maw-other}
In this experiment we studied performance of three different
algorithm: MAW, MAC~\cite{wagner-mac}, and Random Walk. All algorithms
used the same square effector of size $5\times5$ pixels; additionally,
the steps of the Random Walk algorithm were restricted to be in
interval $[r,2r]$ just like the steps in the MAW algorithm.
\begin{figure}[H]%
  \centering
  \subfloat[Cover Time: Domain A]%
  {\label{fig:cover-time-a}%
    \includegraphics[width=\textwidth,height=0.3\textheight]%
    {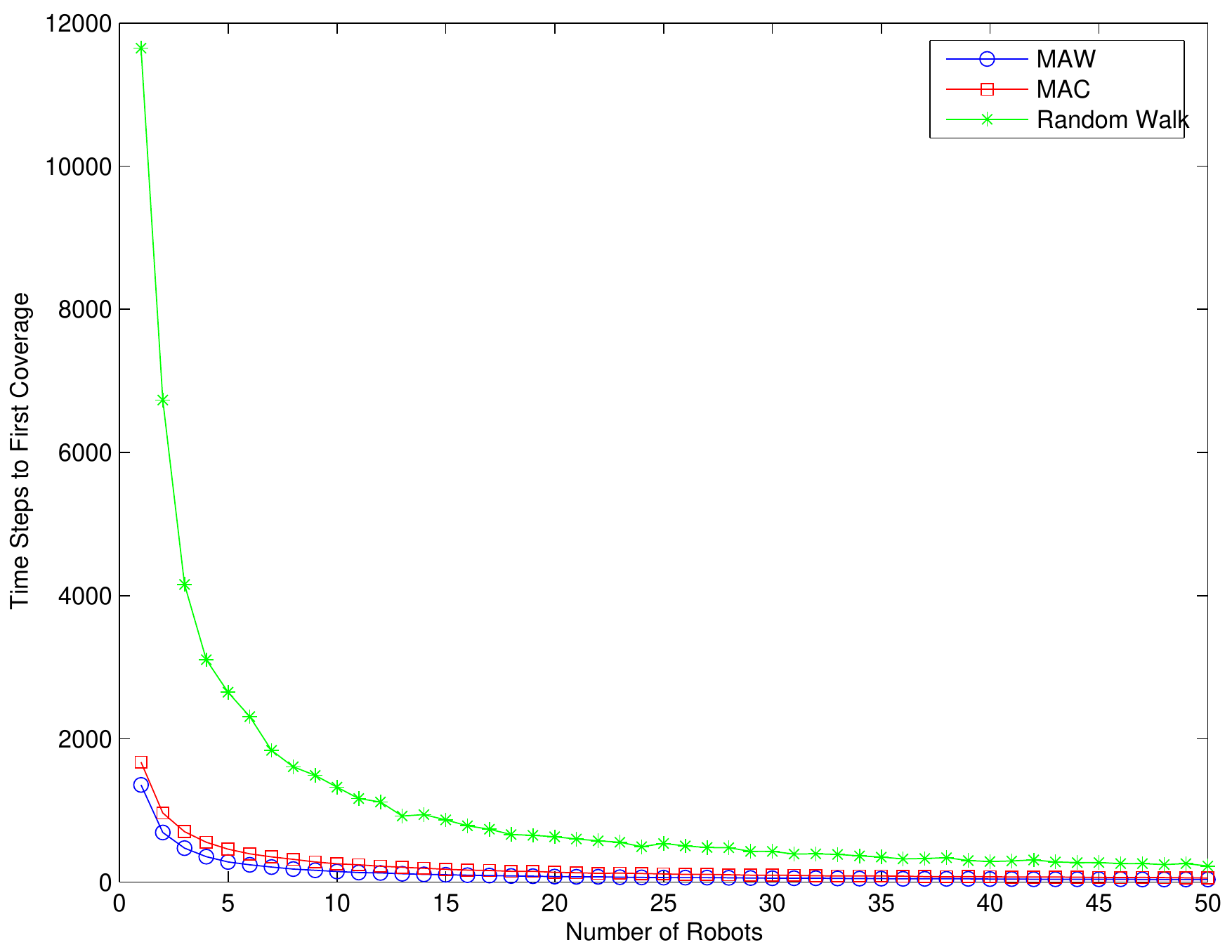}
  }
  \caption{Cover Time}
  \label{fig:cover-time}
\end{figure}
\begin{figure}[H]%
  \ContinuedFloat
  \centering
  \subfloat[Cover Time: Domain B]%
  {\label{fig:cover-time-b}%
    \includegraphics[width=\textwidth,height=0.3\textheight]%
    {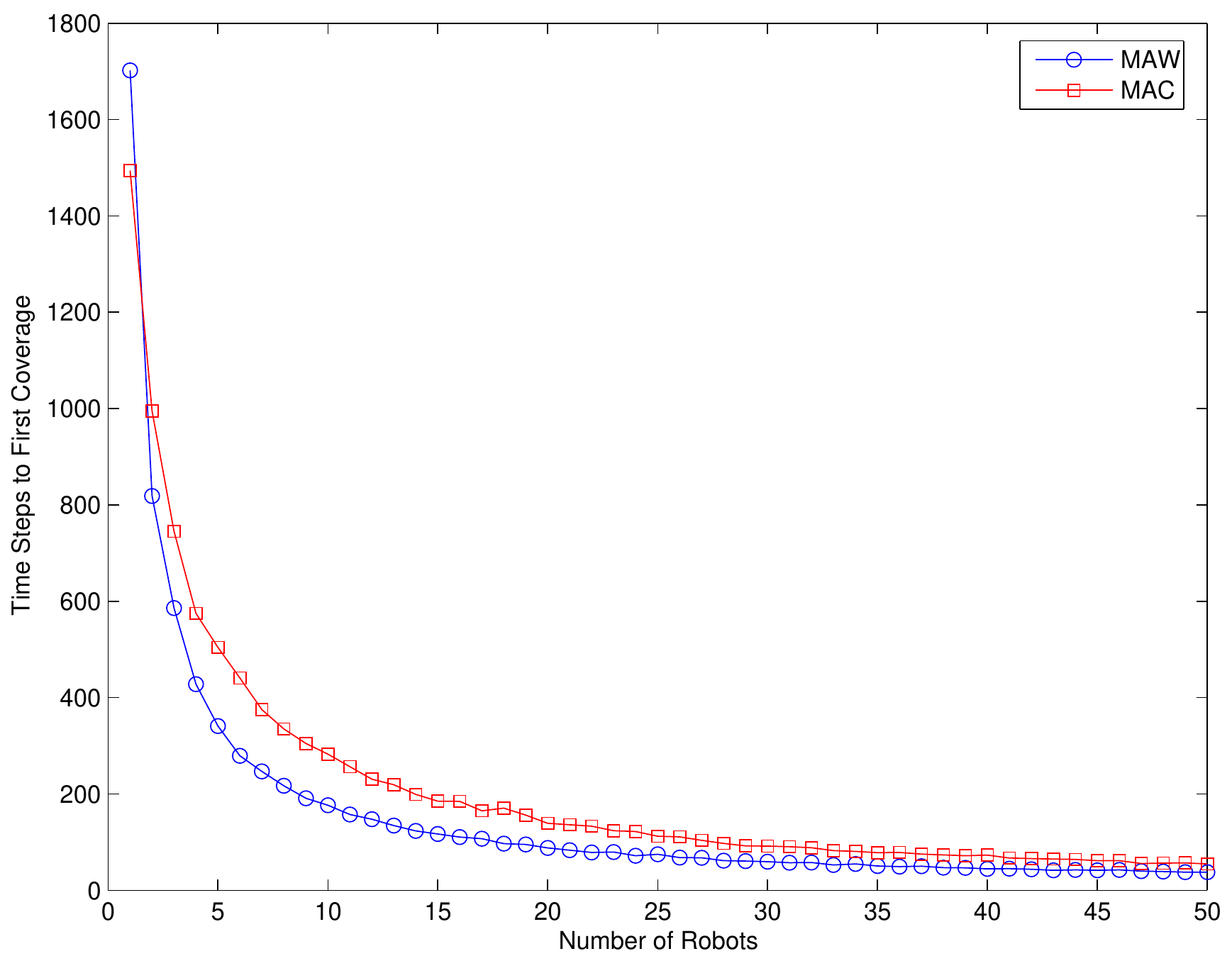} 
  }\\
  \subfloat[Cover Time: Domain C]%
  {\label{fig:cover-time-c}%
    \includegraphics[width=\textwidth,height=0.3\textheight]%
    {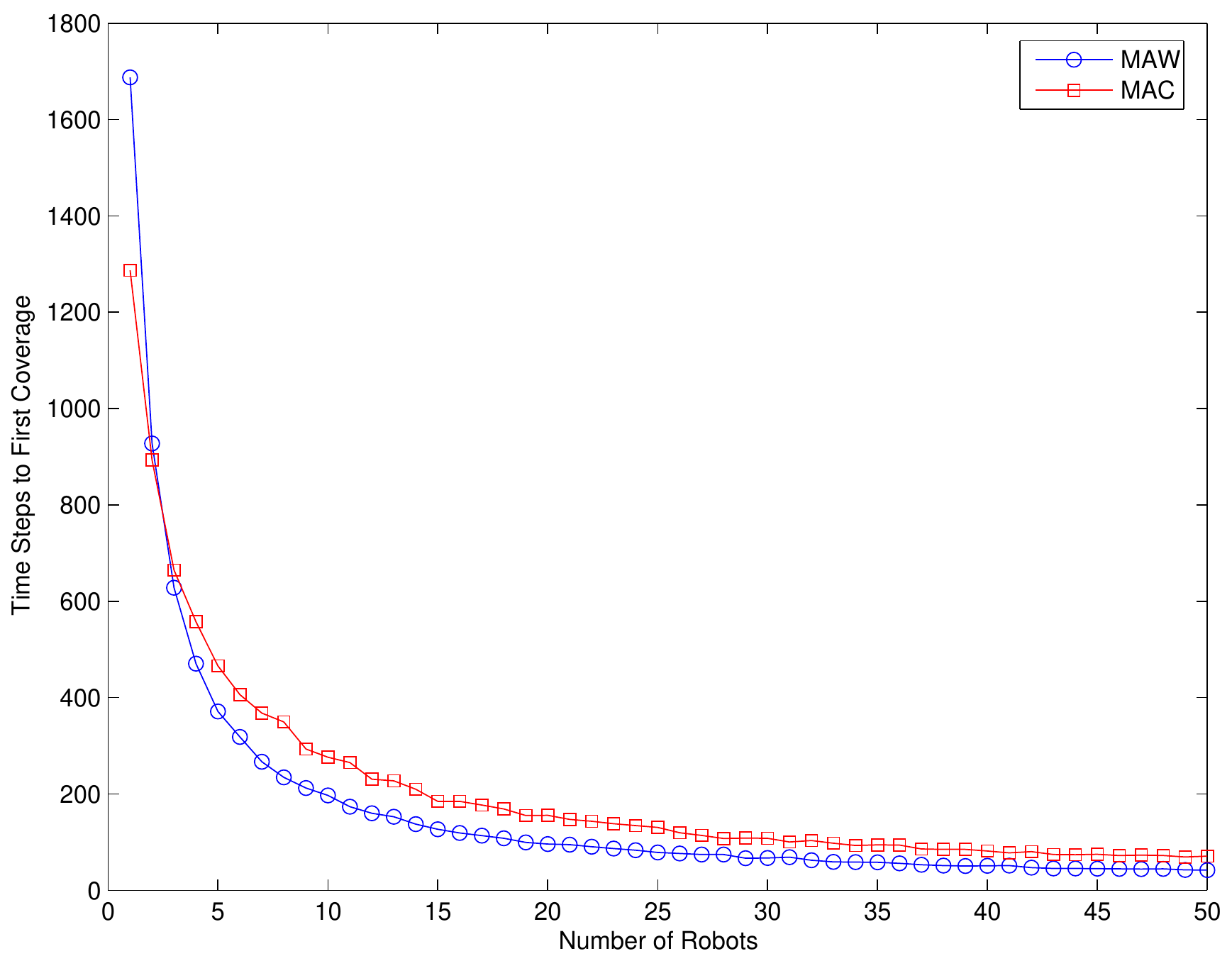} 
  }
  \caption[]{Cover Time}
  \label{fig:cover-time}
\end{figure}

As we can see the MAW algorithm is a clear winner when we use three or
more robots. For fewer robots the MAC algorithm performs better on
complex domains. Note that the MAW algorithm in general performs better than the
theoretical upper bound we got in
Section~\ref{sec:maw-effic-analys}. Cover time of the Random Walk was
omitted from Figures~\ref{fig:cover-time-b} and~\ref{fig:cover-time-c}
because the values were so big that the difference between the MAC and
the MAW algorithms became invisible on this scale. Full results with
additional statistical data can be seen in Appendix~\ref{sec:app_A}.

\begin{figure}[H]
  \centering
  \includegraphics[width=0.8\textwidth]{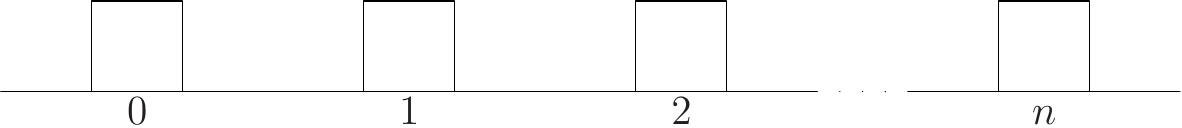}
  \caption{Domain comprised of $n$ loops}
  \label{fig:loops_domain}
\end{figure}

\begin{figure}[H]
  \centering
  \includegraphics[width=0.9\textwidth]{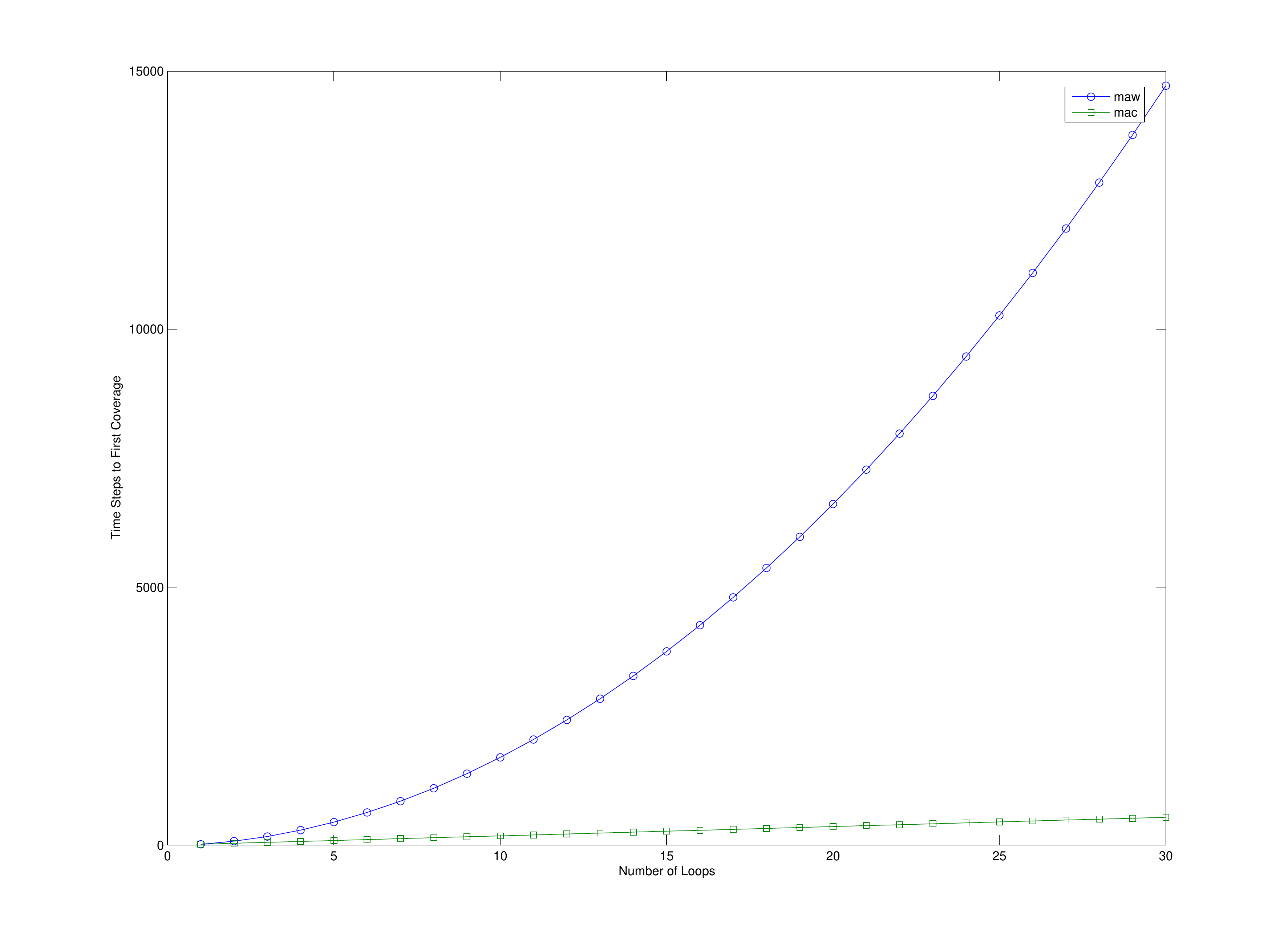}
  \caption{Worst case coverage time of specific domain}
  \label{fig:loops_results}
\end{figure}

Note that our upper bound on coverage time is quadratic, while the
above results suggest that the actual coverage time is linear.
We can demonstrate the predicted quadratic coverage time by tailoring
specific tie breaking rules for specific domain. For example we can
use a domain comprised of $n$ loops as shown in
Figure~\ref{fig:loops_domain}.
For this domain and specific tie breaking rules we can obtain quadratic
coverage time for MAW algorithm, while MAC algorithm still demonstrates
linear coverage time. These results are shown in
Figure~\ref{fig:loops_results}.

\section{MAW in noisy environments}
\label{sec:maw-noisy-envir}
\begin{figure}[H]
  \centering
  \includegraphics[width=\textwidth,height=0.3\textheight]%
  {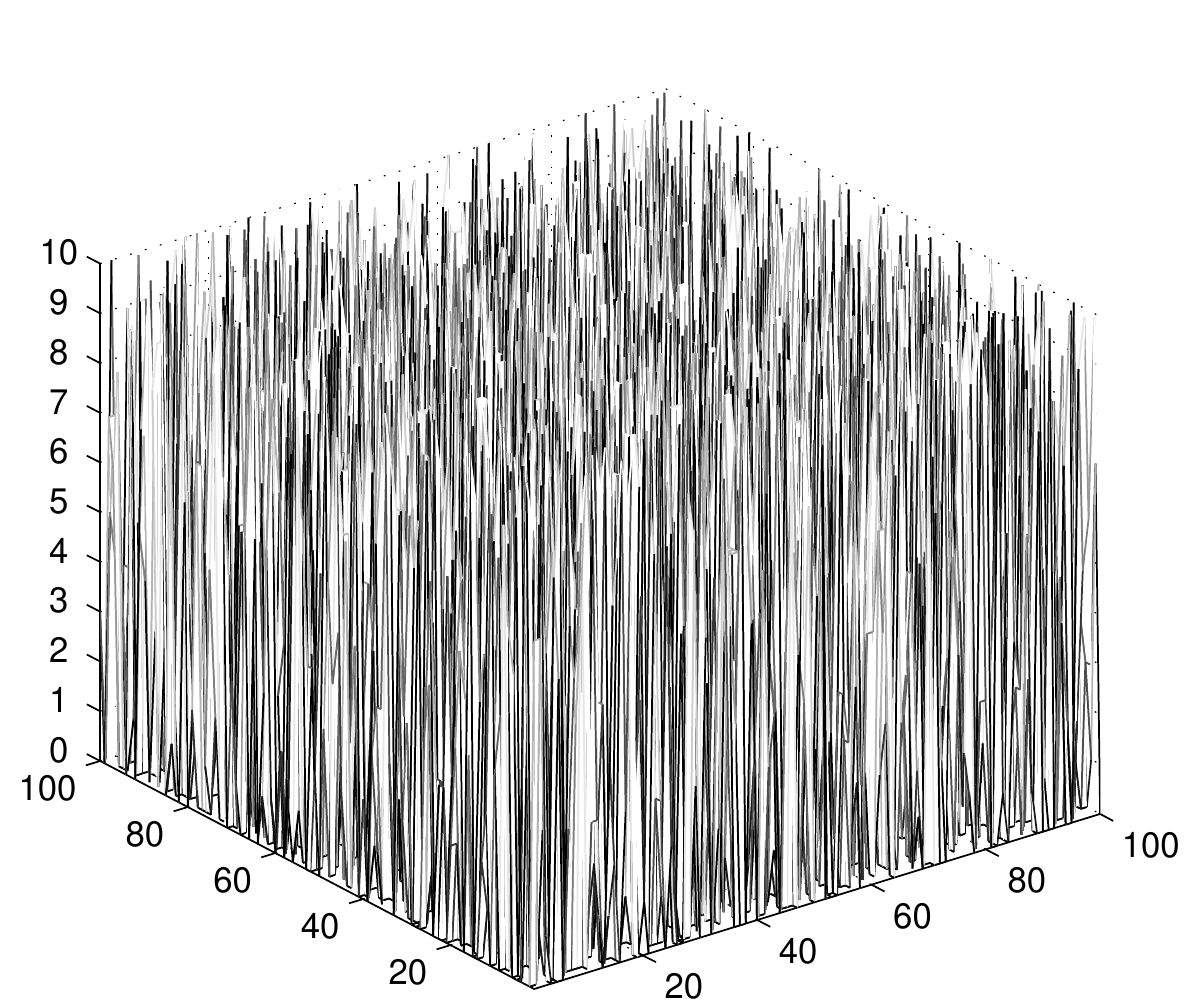} 
  \caption{Example of noise profile 1.}
  \label{fig:noise1}
\end{figure}

\begin{figure}[H]
  \centering
  \includegraphics[width=\textwidth,height=0.3\textheight]%
  {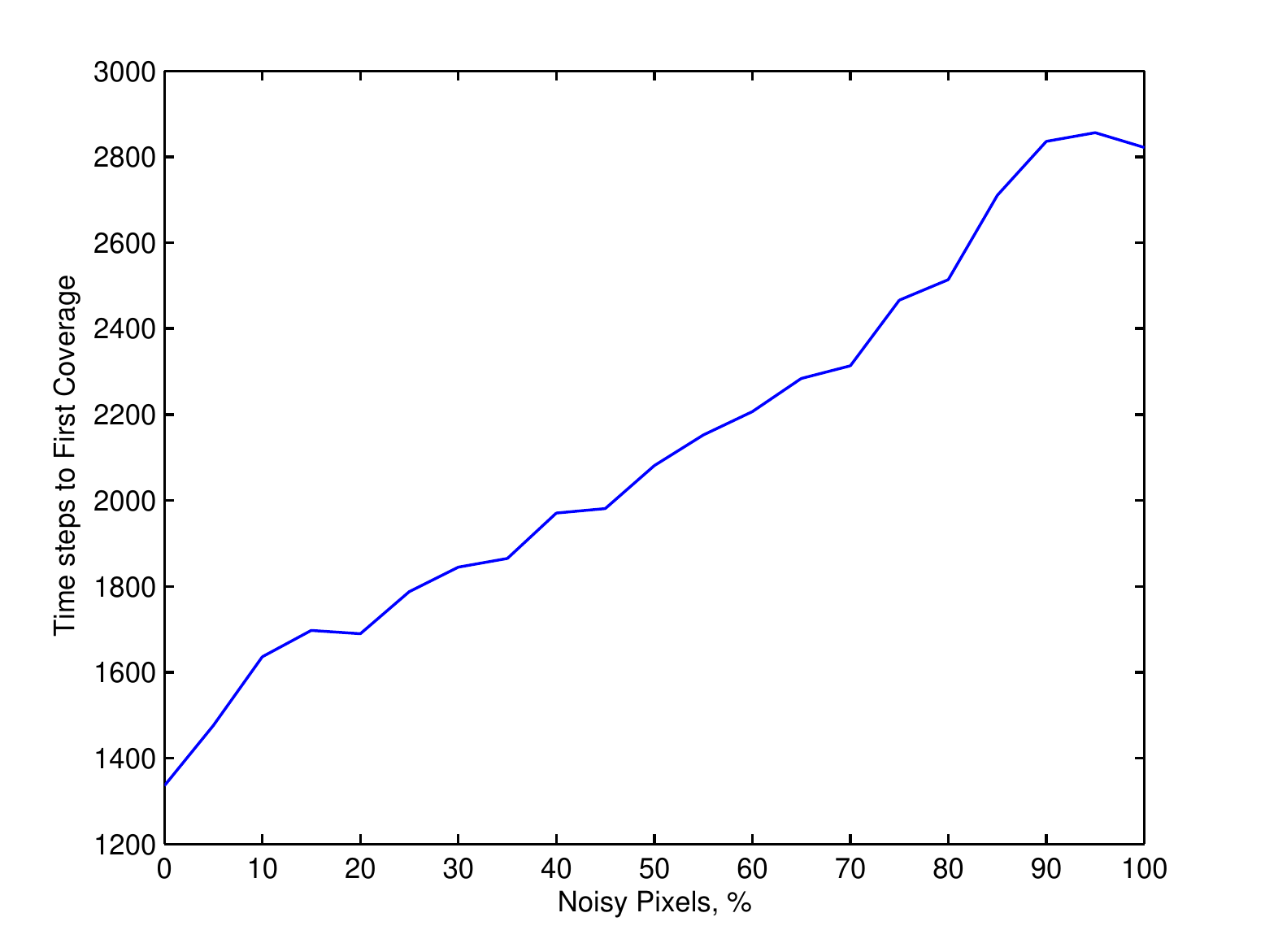} 
  \caption{Cover time in noisy environment}
  \label{fig:noise-time}
\end{figure}

In this section we present the results of our simulation of MAW
algorithm in presence of noise. In the first scenario we ran one robot
on the Domain A, each time changing 
the amount of noisy pixels. Noise values are uniformly distributed in
interval $[1, 10]$, i.e., given that 60 percent of the pixels are
noisy there are about 6 percent that got value of 1, 6 percent
that got value of 2 and so on. Example of such noise profile with
60 percent noisy pixels is shown in Figure~\ref{fig:noise1}.
In Figure~\ref{fig:noise-time} we demonstrate the cover time as a 
function of the amount of noisy pixels in this scenario.

\begin{figure}[H]
  \centering
  \includegraphics[width=\textwidth,height=0.3\textheight]%
  {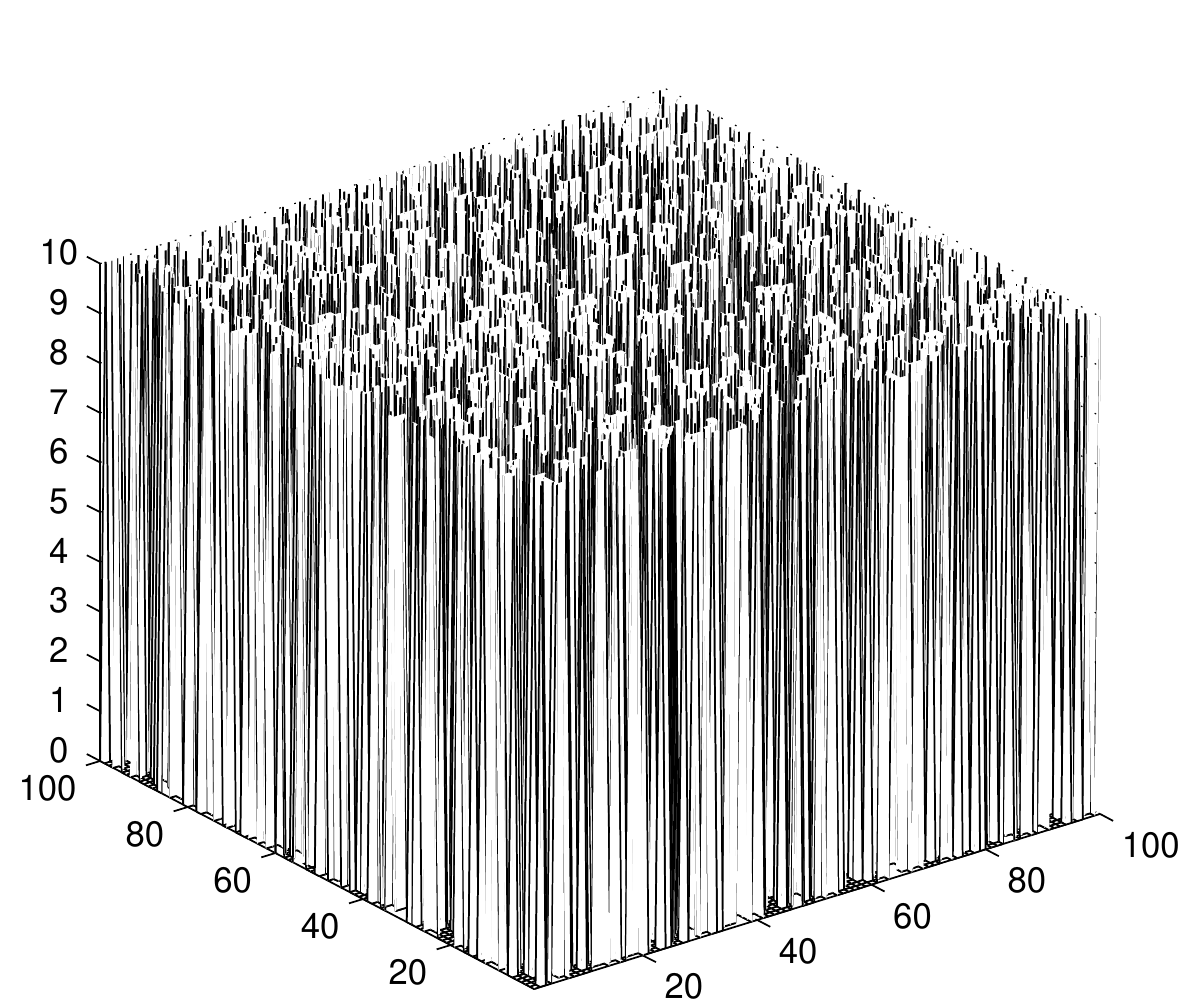} 
  \caption{Example of noise profile 2.}
  \label{fig:noise2}
\end{figure}

\begin{figure}[H]
  \centering
  \includegraphics[width=\textwidth,height=0.3\textheight]%
  {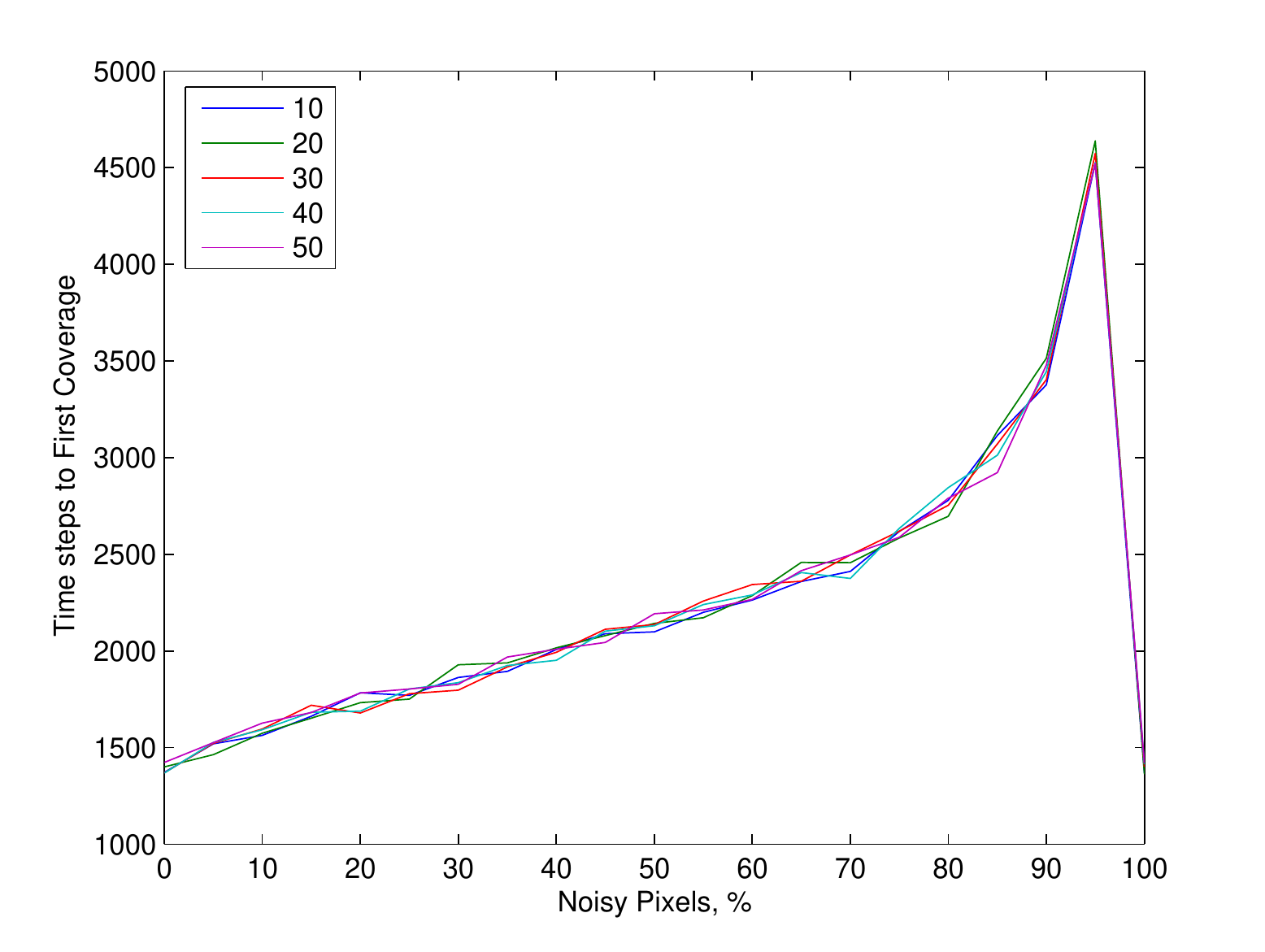} 
  \caption{Cover time in noisy environment}
  \label{fig:noise-time2}
\end{figure}

\begin{figure}[H]
  \centering
  \includegraphics[width=\textwidth,height=0.3\textheight]%
  {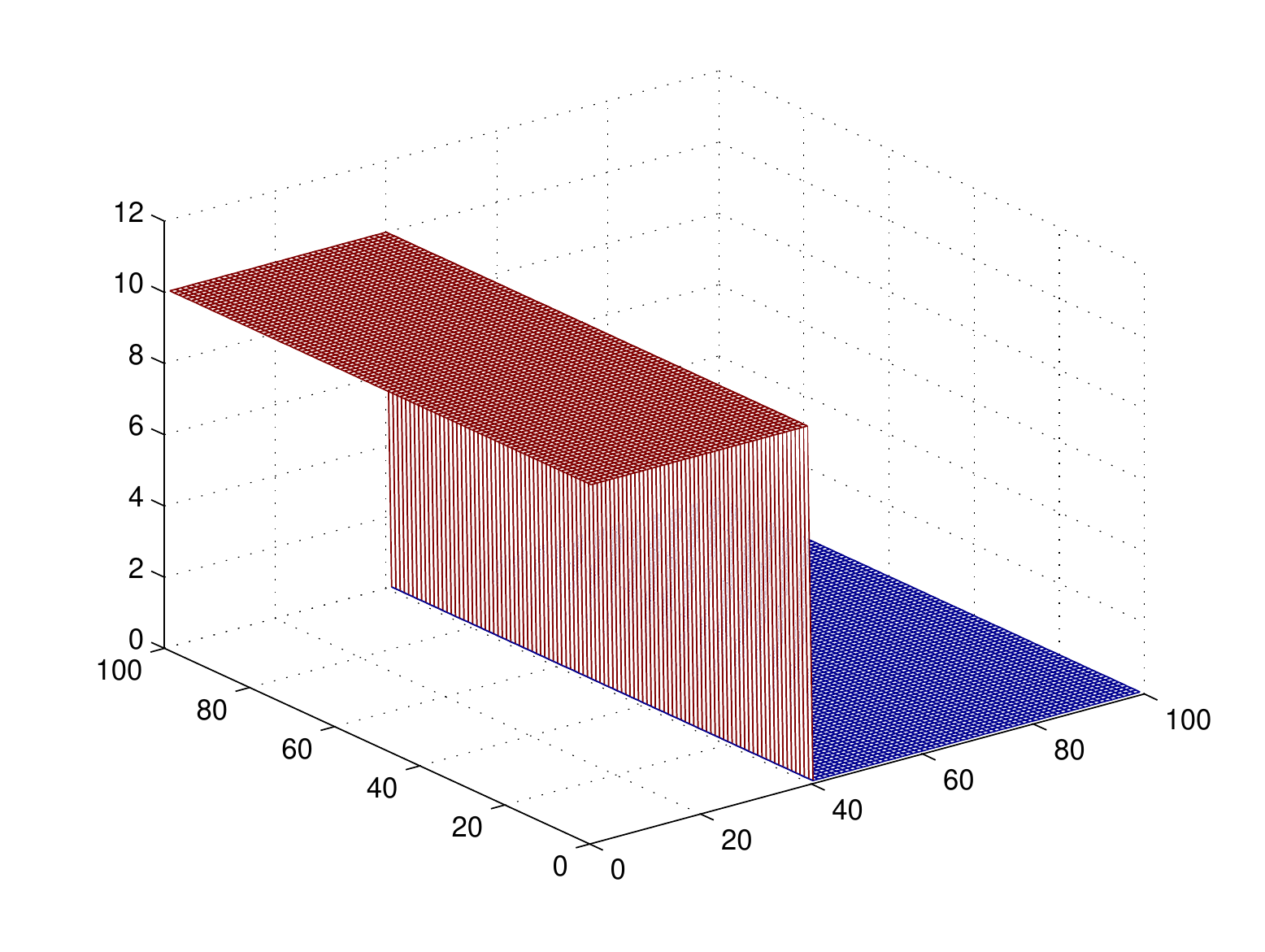} 
  \caption{Example of noise profile 3.}
  \label{fig:noise3}
\end{figure}

\begin{figure}[H]
  \centering
  \includegraphics[width=\textwidth,height=0.3\textheight]%
  {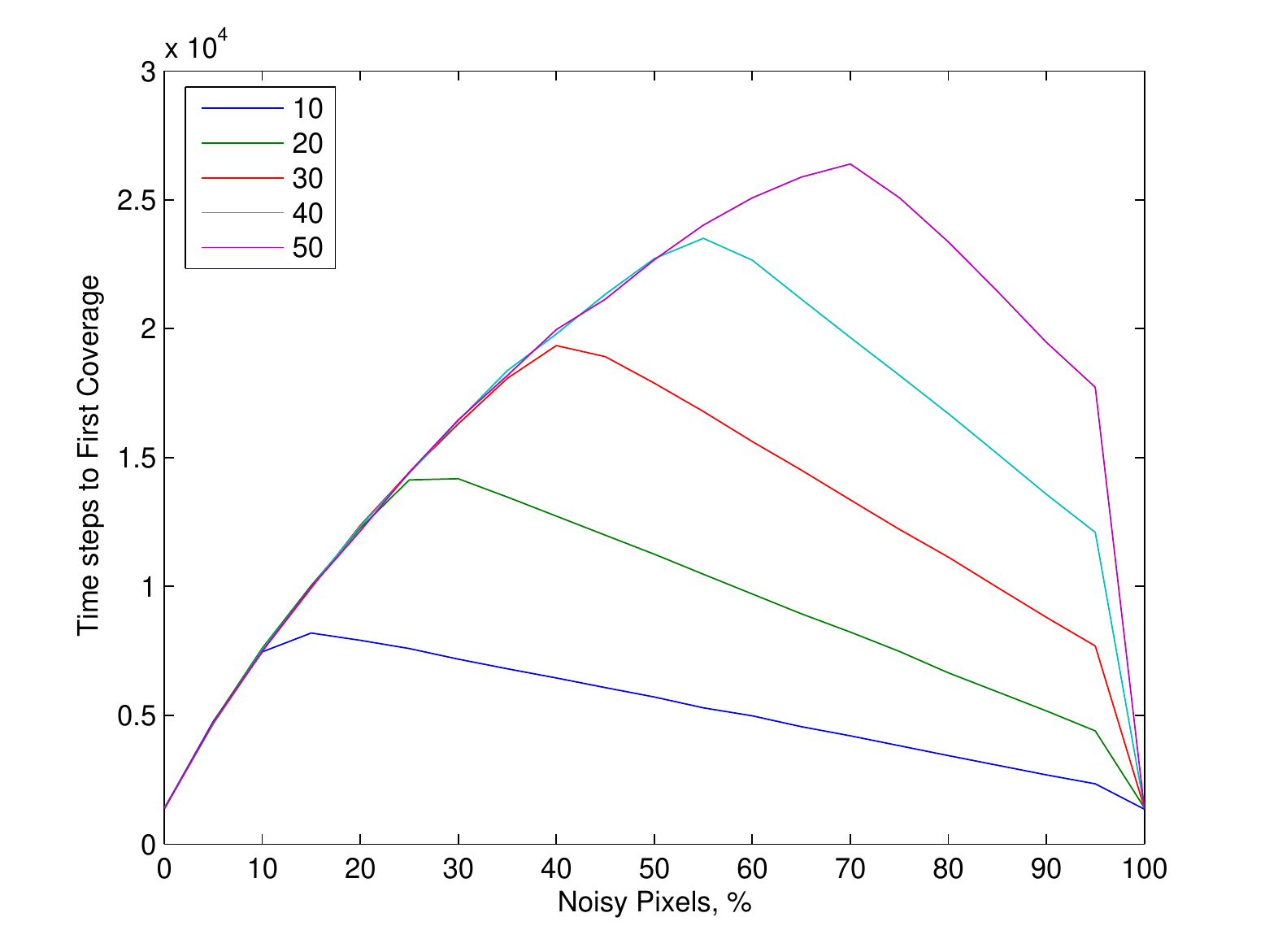} 
  \caption{Cover time in noisy environment}
  \label{fig:noise-time3}
\end{figure}

In another scenario we chose to explore the influence of constant
noise values on the performance of the algorithm. This time noise
values in each experiment were constant and again randomly distributed
in the space. Example of such noise profile for noise value of 10 and
30 percent noisy pixels is shown in Figure~\ref{fig:noise2}.
We run a series of experiments with the noise values of
10, 20, 30, 40, and 50. The results of the cover time versus the
percentage of noisy pixels is shown in Figure~\ref{fig:noise-time2}. 
As you can see the value of the noise does not play any role in this
scenario, at least in the limits we used here, from 10 to 50. This is
probably due to the nature of the algorithm that knows to discard
high pheromone values in presence of lower values. To check this we
conducted another experiment, that is similar to this one, however
noise is this scenario occupies a compact space in the domain, i.e.,
given that there 40 percent of noisy pixels we form a plateau of noise
as shown in Figure~\ref{fig:noise3}
In this case the influence of the noise value if pronounced as one can
expect. See Figure~\ref{fig:noise-time3} for covering time versus
noisy pixel percentage for noise values of 10, 20, 30, 40, and 50.
As our experiments demonstrate the MAW algorithm has little
sensitivity no ``non-compact'' distribution of the noise. 
Note that noise does not affect the Random Walk on the one hand and it
destroys completely the MAC algorithm on the other hand, making it
unable to cover the domain completely.
\chapter{Conclusions}
\label{sec:conclusions}
This work has two major contributions. First,  we presented a new
ant-inspired algorithm for continuous 
domain covering. We provided a formal proof of complete coverage
and upper time bounds for complete coverage and the time interval
between two successive visits of the robot. We also proved that the
algorithm is immune to pheromone noise in the  environment. A formal
proof was provided for multi-robot coverage under the assumptions that the
robots have different clock phases.
Second, a new way of  performance estimation was suggested, that implies
some bounds on possible coverage time of any algorithm. 
% \section{Future Work}
% \label{sec:future-work}
% \begin{itemize}
% \item show that the coverage time bound is tight
% \item show that repetitive coverage time bound is tight
% \item dynamic noise
% \item multiple robots (better analysis)
% \end{itemize}

\appendix
\chapter{Detailed Statistics of MAW, MAC, and PC performance}
\label{sec:app_A}

\begin{table}[ht]
  \caption{MAW Results on Domain A}
  \centering
  \begin{tabular}{|c|c|c|c|c|}
\hline
\# robots & mean & max & min & std\\
\hline
1 & 1372.2 & 1999 & 1122 & 169.4\\
\hline
2 & 696.6 & 888 & 550 & 76.7\\
\hline
3 & 472.4 & 647 & 352 & 55.5\\
\hline
4 & 357.1 & 510 & 276 & 46.8\\
\hline
5 & 285.9 & 410 & 225 & 33.0\\
\hline
6 & 240.5 & 337 & 195 & 26.3\\
\hline
7 & 208.1 & 294 & 154 & 26.6\\
\hline
8 & 180.8 & 243 & 139 & 20.4\\
\hline
9 & 160.8 & 223 & 125 & 19.2\\
\hline
10 & 146.3 & 203 & 115 & 18.4\\
\hline
11 & 133.5 & 182 & 103 & 17.4\\
\hline
12 & 124.6 & 170 & 96 & 16.3\\
\hline
13 & 113.7 & 155 & 89 & 13.7\\
\hline
14 & 107.4 & 146 & 82 & 13.5\\
\hline
15 & 101.7 & 149 & 76 & 12.5\\
\hline
16 & 93.1 & 130 & 73 & 10.7\\
\hline
17 & 89.1 & 119 & 71 & 10.0\\
\hline
18 & 84.2 & 110 & 66 & 9.3\\
\hline
19 & 79.2 & 103 & 64 & 8.8\\
\hline
20 & 76.8 & 114 & 60 & 10.3\\
\hline
21 & 71.2 & 90 & 58 & 6.7\\
\hline
22 & 70.9 & 102 & 56 & 10.0\\
\hline
23 & 66.9 & 101 & 52 & 9.0\\
\hline
24 & 65.6 & 83 & 50 & 7.4\\
\hline
25 & 61.5 & 88 & 50 & 7.0\\
\hline
26 & 59.5 & 82 & 48 & 7.6\\
\hline
27 & 59.1 & 86 & 46 & 7.6\\
\hline
28 & 57.4 & 80 & 44 & 7.2\\
\hline
29 & 55.0 & 73 & 44 & 6.5\\
\hline
30 & 52.9 & 78 & 41 & 7.0\\
\hline
31 & 50.9 & 80 & 41 & 6.1\\
\hline
32 & 49.7 & 64 & 38 & 4.9\\
\hline
33 & 50.0 & 62 & 39 & 6.2\\
\hline
34 & 47.9 & 67 & 38 & 5.6\\
\hline
35 & 46.8 & 72 & 37 & 6.7\\
\hline
\end{tabular}

  \label{tab:maw_A}
\end{table}

\begin{table}[ht]
  \caption{MAW Results on Domain B}
  \centering
  \begin{tabular}{|c|c|c|c|c|}
\hline
\# robots & mean & max & min & std\\
\hline
1 & 1689.6 & 3878 & 1021 & 500.8\\
\hline
2 & 830.5 & 1690 & 501 & 226.7\\
\hline
3 & 582.8 & 1411 & 365 & 193.9\\
\hline
4 & 454.6 & 1032 & 279 & 150.4\\
\hline
5 & 362.5 & 930 & 206 & 116.8\\
\hline
6 & 288.6 & 584 & 190 & 72.7\\
\hline
7 & 269.5 & 643 & 170 & 95.1\\
\hline
8 & 204.5 & 559 & 132 & 58.9\\
\hline
9 & 192.0 & 429 & 117 & 54.6\\
\hline
10 & 174.8 & 349 & 112 & 49.7\\
\hline
11 & 155.6 & 292 & 103 & 41.9\\
\hline
12 & 144.7 & 305 & 93 & 41.5\\
\hline
13 & 127.3 & 263 & 88 & 34.3\\
\hline
14 & 123.6 & 264 & 73 & 33.2\\
\hline
15 & 116.7 & 248 & 73 & 31.4\\
\hline
16 & 112.9 & 231 & 69 & 31.9\\
\hline
17 & 102.3 & 227 & 66 & 26.4\\
\hline
18 & 98.2 & 204 & 64 & 25.7\\
\hline
19 & 95.5 & 207 & 63 & 26.3\\
\hline
20 & 91.5 & 199 & 57 & 25.1\\
\hline
21 & 84.8 & 153 & 60 & 18.9\\
\hline
22 & 80.1 & 143 & 54 & 18.0\\
\hline
23 & 76.3 & 149 & 50 & 18.3\\
\hline
24 & 75.6 & 177 & 50 & 23.3\\
\hline
25 & 69.8 & 185 & 48 & 20.0\\
\hline
26 & 69.3 & 147 & 43 & 19.0\\
\hline
27 & 65.1 & 127 & 44 & 14.6\\
\hline
28 & 63.9 & 153 & 44 & 16.8\\
\hline
29 & 63.2 & 172 & 40 & 19.2\\
\hline
30 & 59.4 & 111 & 39 & 13.8\\
\hline
31 & 57.0 & 106 & 40 & 12.5\\
\hline
32 & 56.8 & 105 & 36 & 13.1\\
\hline
33 & 53.1 & 93 & 35 & 10.5\\
\hline
34 & 53.9 & 102 & 34 & 12.7\\
\hline
35 & 49.0 & 77 & 34 & 8.6\\
\hline
\end{tabular}

  \label{tab:maw_B}
\end{table}

\begin{table}[ht]
  \caption{MAW Results on Domain C}
  \centering
  \begin{tabular}{|c|c|c|c|c|}
\hline
\# robots & mean & max & min & std\\
\hline
1 & 1713.3 & 4143 & 1076 & 579.0\\
\hline
2 & 881.0 & 2413 & 492 & 355.3\\
\hline
3 & 558.3 & 1176 & 342 & 149.5\\
\hline
4 & 444.7 & 1009 & 258 & 130.0\\
\hline
5 & 335.3 & 650 & 222 & 89.6\\
\hline
6 & 294.4 & 674 & 191 & 91.5\\
\hline
7 & 243.8 & 579 & 162 & 69.4\\
\hline
8 & 227.0 & 698 & 121 & 83.9\\
\hline
9 & 190.9 & 421 & 124 & 51.2\\
\hline
10 & 176.1 & 446 & 115 & 54.6\\
\hline
11 & 158.2 & 356 & 103 & 49.4\\
\hline
12 & 148.8 & 332 & 88 & 40.1\\
\hline
13 & 133.6 & 251 & 87 & 35.7\\
\hline
14 & 129.5 & 326 & 84 & 39.7\\
\hline
15 & 118.2 & 216 & 75 & 29.8\\
\hline
16 & 110.8 & 202 & 70 & 29.5\\
\hline
17 & 99.9 & 196 & 60 & 23.9\\
\hline
18 & 96.0 & 175 & 69 & 20.5\\
\hline
19 & 92.8 & 167 & 67 & 23.9\\
\hline
20 & 89.7 & 195 & 56 & 23.2\\
\hline
21 & 82.2 & 193 & 52 & 21.2\\
\hline
22 & 77.1 & 131 & 53 & 14.9\\
\hline
23 & 78.4 & 227 & 51 & 25.4\\
\hline
24 & 77.2 & 203 & 49 & 21.6\\
\hline
25 & 72.5 & 166 & 51 & 18.6\\
\hline
26 & 66.5 & 117 & 46 & 12.0\\
\hline
27 & 65.7 & 104 & 40 & 13.2\\
\hline
28 & 64.8 & 130 & 42 & 16.6\\
\hline
29 & 61.1 & 97 & 39 & 11.9\\
\hline
30 & 60.1 & 89 & 40 & 11.5\\
\hline
31 & 56.5 & 109 & 35 & 14.1\\
\hline
32 & 57.2 & 102 & 39 & 13.4\\
\hline
33 & 55.2 & 94 & 39 & 12.4\\
\hline
34 & 50.7 & 96 & 35 & 9.2\\
\hline
35 & 52.1 & 117 & 35 & 12.0\\
\hline
\end{tabular}

  \label{tab:maw_C}
\end{table}

\begin{table}[ht]
  \caption{MAC Results on Domain A}
  \centering
  \begin{tabular}{|c|c|c|c|c|}
\hline
\# robots & mean & max & min & std\\
\hline
1 & 1679.7 & 1728 & 1586 & 26.9\\
\hline
2 & 966.1 & 1503 & 807 & 138.9\\
\hline
3 & 686.5 & 1078 & 540 & 99.1\\
\hline
4 & 565.2 & 885 & 423 & 92.2\\
\hline
5 & 451.6 & 734 & 318 & 67.9\\
\hline
6 & 386.3 & 710 & 300 & 67.8\\
\hline
7 & 347.9 & 551 & 244 & 66.0\\
\hline
8 & 302.7 & 655 & 222 & 55.5\\
\hline
9 & 275.1 & 444 & 196 & 53.4\\
\hline
10 & 260.0 & 495 & 184 & 53.8\\
\hline
11 & 234.8 & 439 & 167 & 47.8\\
\hline
12 & 214.8 & 338 & 154 & 37.6\\
\hline
13 & 198.3 & 276 & 144 & 30.8\\
\hline
14 & 188.9 & 304 & 131 & 39.4\\
\hline
15 & 179.2 & 282 & 123 & 34.9\\
\hline
16 & 166.7 & 286 & 110 & 35.6\\
\hline
17 & 159.9 & 244 & 110 & 26.7\\
\hline
18 & 148.7 & 237 & 107 & 27.6\\
\hline
19 & 142.0 & 226 & 95 & 26.7\\
\hline
20 & 140.6 & 322 & 90 & 34.7\\
\hline
21 & 135.8 & 299 & 86 & 35.2\\
\hline
22 & 126.3 & 226 & 87 & 24.0\\
\hline
23 & 119.6 & 195 & 86 & 22.2\\
\hline
24 & 117.8 & 206 & 76 & 27.0\\
\hline
25 & 114.0 & 202 & 76 & 26.5\\
\hline
26 & 105.1 & 148 & 69 & 18.8\\
\hline
27 & 103.8 & 190 & 72 & 19.2\\
\hline
28 & 104.3 & 170 & 70 & 22.2\\
\hline
29 & 95.5 & 139 & 69 & 15.4\\
\hline
30 & 95.0 & 182 & 63 & 20.1\\
\hline
31 & 94.6 & 170 & 65 & 18.3\\
\hline
32 & 89.8 & 155 & 61 & 20.2\\
\hline
33 & 88.0 & 181 & 61 & 18.5\\
\hline
34 & 81.4 & 125 & 53 & 14.7\\
\hline
35 & 80.0 & 135 & 50 & 15.7\\
\hline
\end{tabular}

  \label{tab:mac_A}
\end{table}

\begin{table}[ht]
  \caption{MAC Results on Domain B}
  \centering
  \begin{tabular}{|c|c|c|c|c|}
\hline
\# robots & mean & max & min & std\\
\hline
1 & 1491.2 & 1540 & 1389 & 26.4\\
\hline
2 & 986.1 & 1344 & 723 & 172.4\\
\hline
3 & 729.4 & 1238 & 501 & 165.0\\
\hline
4 & 601.7 & 1090 & 374 & 150.3\\
\hline
5 & 479.9 & 807 & 300 & 117.3\\
\hline
6 & 410.6 & 716 & 266 & 92.1\\
\hline
7 & 377.4 & 723 & 223 & 108.7\\
\hline
8 & 352.6 & 719 & 206 & 106.2\\
\hline
9 & 304.2 & 630 & 168 & 94.8\\
\hline
10 & 290.7 & 558 & 171 & 89.4\\
\hline
11 & 262.3 & 525 & 147 & 86.1\\
\hline
12 & 238.2 & 485 & 127 & 74.9\\
\hline
13 & 225.8 & 500 & 128 & 76.0\\
\hline
14 & 208.2 & 440 & 115 & 67.1\\
\hline
15 & 201.2 & 414 & 114 & 69.1\\
\hline
16 & 173.3 & 412 & 91 & 52.7\\
\hline
17 & 163.1 & 373 & 96 & 48.8\\
\hline
18 & 163.1 & 390 & 100 & 55.9\\
\hline
19 & 147.5 & 274 & 92 & 43.4\\
\hline
20 & 146.6 & 367 & 86 & 50.2\\
\hline
21 & 133.1 & 338 & 85 & 46.0\\
\hline
22 & 125.0 & 332 & 84 & 35.5\\
\hline
23 & 123.1 & 372 & 77 & 37.6\\
\hline
24 & 118.1 & 243 & 70 & 31.6\\
\hline
25 & 121.7 & 322 & 71 & 42.8\\
\hline
26 & 104.0 & 218 & 63 & 28.2\\
\hline
27 & 103.4 & 209 & 66 & 26.1\\
\hline
28 & 95.3 & 164 & 68 & 21.2\\
\hline
29 & 96.8 & 322 & 64 & 32.7\\
\hline
30 & 94.3 & 186 & 61 & 24.5\\
\hline
31 & 88.0 & 164 & 59 & 21.2\\
\hline
32 & 89.6 & 257 & 56 & 26.7\\
\hline
33 & 81.5 & 195 & 56 & 21.0\\
\hline
34 & 80.2 & 144 & 48 & 17.4\\
\hline
35 & 79.0 & 146 & 54 & 17.4\\
\hline
\end{tabular}

  \label{tab:mac_B}
\end{table}

\begin{table}[ht]
  \caption{MAC Results on Domain C}
  \centering
  \begin{tabular}{|c|c|c|c|c|}
\hline
\# robots & mean & max & min & std\\
\hline
1 & 1493.8 & 1559 & 1429 & 24.4\\
\hline
2 & 975.0 & 1361 & 752 & 175.0\\
\hline
3 & 705.2 & 1133 & 484 & 123.4\\
\hline
4 & 577.3 & 952 & 358 & 112.9\\
\hline
5 & 485.6 & 712 & 296 & 101.5\\
\hline
6 & 410.1 & 705 & 254 & 102.7\\
\hline
7 & 392.8 & 652 & 222 & 103.2\\
\hline
8 & 334.9 & 594 & 213 & 86.2\\
\hline
9 & 294.7 & 707 & 181 & 90.0\\
\hline
10 & 274.3 & 558 & 144 & 75.6\\
\hline
11 & 261.5 & 548 & 151 & 87.1\\
\hline
12 & 229.6 & 481 & 145 & 65.5\\
\hline
13 & 226.2 & 473 & 125 & 73.6\\
\hline
14 & 200.2 & 512 & 119 & 61.6\\
\hline
15 & 185.9 & 375 & 108 & 56.3\\
\hline
16 & 165.9 & 361 & 111 & 45.5\\
\hline
17 & 164.9 & 428 & 95 & 53.8\\
\hline
18 & 160.8 & 380 & 84 & 53.6\\
\hline
19 & 146.3 & 311 & 94 & 42.7\\
\hline
20 & 144.4 & 290 & 86 & 40.1\\
\hline
21 & 132.6 & 320 & 85 & 39.0\\
\hline
22 & 127.6 & 262 & 70 & 35.2\\
\hline
23 & 118.1 & 390 & 75 & 39.5\\
\hline
24 & 114.4 & 218 & 77 & 29.8\\
\hline
25 & 110.1 & 288 & 74 & 36.5\\
\hline
26 & 108.9 & 240 & 65 & 31.5\\
\hline
27 & 103.6 & 258 & 67 & 32.7\\
\hline
28 & 100.4 & 193 & 65 & 26.6\\
\hline
29 & 96.2 & 171 & 64 & 23.7\\
\hline
30 & 94.7 & 176 & 59 & 22.7\\
\hline
31 & 91.5 & 170 & 59 & 21.6\\
\hline
32 & 86.7 & 178 & 59 & 22.8\\
\hline
33 & 84.2 & 188 & 55 & 22.1\\
\hline
34 & 81.9 & 170 & 52 & 18.9\\
\hline
35 & 79.5 & 173 & 55 & 19.6\\
\hline
\end{tabular}

  \label{tab:mac_C}
\end{table}

\bibliographystyle{chicago}
\bibliography{biblio}

% \begin{otherlanguage}{hebrew}
%   \thesisheaderminor

%   \begin{abstract}
%     \input{hebabstract}
%   \end{abstract}
% \end{otherlanguage}

\end{document}